\documentclass[journal]{IEEEtran}

\usepackage{amssymb}
\usepackage{amsmath}
\usepackage{graphicx}
\usepackage{bbold}
\usepackage{xcolor}
\usepackage{multirow}
\usepackage{amsmath}
\renewcommand{\baselinestretch}{1.0}

\usepackage{url}
\urldef{\mailsa}\path|{alfred.hofmann, ursula.barth, ingrid.haas, frank.holzwarth,|
\urldef{\mailsb}\path|anna.kramer, leonie.kunz, christine.reiss, nicole.sator,|
\urldef{\mailsc}\path|erika.siebert-cole, petcaner.strasser, lncs}@springer.com|

\newcommand{\etal}{\emph{et al.}~}
\newcommand{\ie}{\emph{i.e.}}
\newcommand{\eg}{\emph{e.g.}}

\begin{document}

\title{On Modelling Label Uncertainty in Deep Neural Networks: Automatic Estimation of Intra-observer Variability in 2D Echocardiography Quality Assessment}

\author{Zhibin Liao, 
Hany Girgis, 
Amir Abdi, 
Hooman Vaseli,
Jorden Hetherington,
Robert~Rohling,
Ken Gin,
Teresa~Tsang,
and Purang~Abolmaesumi
\thanks{Z. Liao and H. Girgis have contributed equally to this work.}
\thanks{The corresponding authors T. Tsang and P. Abolmaesumi have contributed equally to the manuscript (emails: t.tsang@ubc.ca, purang@ece.ubc.ca).}\thanks{Z. Liao, A. Abdi, H. Vaseli, and J. Hetherington are with the Department of Electrical and Computer Engineering, The University of British Columbia, Vancouver, BC V6T 1Z4, Canada.} 
\thanks{H. Girgis, T. Tsang, and K. Gin are with Vancouver General Hospital Echocardiography Laboratory, Division of Cardiology, Department of Medicine, The University of British Columbia, Vancouver, BC V5Z 1M9, Canada.} 
\thanks{R. Rohling is with the Department of Electrical and Computer Engineering and the Department of Mechanical Engineering,
The University of British Columbia, Vancouver, BC V6T 1Z4, Canada}
\thanks{T. Tsang is the Director of the Vancouver General Hospital and University of British Columbia Echocardiography Laboratories, and Principal Investigator of the CIHR-NSERC grant supporting this work.}
\thanks{P. Abolmaesumi is Co-Principal Investigator for the grant supporting this work and is with the Department of Electrical and Computer Engineering, The University of British Columbia, Vancouver, BC V6T 1Z4, Canada.}}

\maketitle

\begin{abstract}
Uncertainty of labels in clinical data resulting from intra-observer variability can have direct impact on the reliability of assessments made by deep neural networks. In this paper, we propose a method for modelling such uncertainty in the context of 2D echocardiography (echo), which is a routine procedure for detecting cardiovascular disease at point-of-care. Echo imaging quality and acquisition time is highly dependent on the operator's experience level. Recent developments have shown the possibility of automating echo image quality quantification by mapping an expert's assessment of quality to the echo image via deep learning techniques. Nevertheless, the observer variability in the expert's assessment can impact the quality quantification accuracy. Here, we aim to model the intra-observer variability in echo quality assessment as an aleatoric uncertainty modelling regression problem with the introduction of a novel method that handles the regression problem with categorical labels. A key feature of our design is that only a single forward pass is sufficient to estimate the level of uncertainty for the network output.
Compared to the $0.11 \pm 0.09$ absolute error (in a scale from 0 to 1) archived by the conventional regression method, the proposed method brings the error down to $0.09 \pm 0.08$, where the improvement is statistically significant and equivalents to 5.7\% test accuracy improvement. The simplicity of the proposed approach means that it could be generalized to other applications of deep learning in medical imaging, where there is often uncertainty in clinical labels.
\end{abstract}


\section{Introduction}
\label{sec:intro}

Medical imaging technologies (\emph{e.g.}, 2D/3D ultrasound (US), Computed Tomography (CT), Magnetic Resonance (MR) imaging) are widely used for screening and diagnosis of cardiovascular disease.
Among these systems, 2D echocardiography (echo) is the primary point-of-care imaging modality for early diagnosis, since it is inexpensive, non-invasive, and widely available.
Accurate acquisition of echo images requires many years of practice to gain adequate anatomical knowledge and hand-eye coordination to adjust imaging to individual patients~\cite{ehler2001guidelines}. 
This means poor quality echo images acquired by less experienced ultrasound operators can negatively affect diagnostic accuracy~\cite{grossgasteiger2014image}.

Beyond conventional methods in estimating ultrasound image quality~\cite{tsantis2014multiresolution, loizou2006quality,huang2014detection, el2013learning, chatelain2016confidence}, most recent methods are based on deep learning to assess image quality.
Wu \etal~\cite{wu2017fuiqa} proposed a convolution neural networks (CNN) for fetal US quality assessment scheme in order to control US image quality.
Abdi \etal showed that deep models can leverage CNN and recurrent neural networks (RNN) to automate the echo quality assessment of images and cine series by learning from experts' labelled samples~\cite{abdi2017automatic,abdi2017quality}.
Although deep learning classifiers are powerful modelling tools, direct mapping from US images to expert labels can be difficult due to observer variability.
In clinical  
studies~\cite{brennan1992statistical, van2013automated}, the lack of consistency in diagnostic judgment and decision making has been long observed.
The variability in clinical labels mainly comes from two sources: 1) the lack of consistency within an observer (\ie, intra-observer variability), and 2) the lack of consistency among observers (\ie, inter-observer variability).

In the machine learning field, one way to treat observer variability is to consider it as label noise problem~\cite{frenay2014classification}.
Data cleaning methods have been used to identify and clean noise samples before training a classifier~\cite{brodley1999identifying, miranda2009use}, but hard (informative but difficult for a model to predict the correct label) samples may also be removed as they can be confused with random noise. 
The noise-robust and noise-tolerant loss functions and models~\cite{beigman2009learning, bartlett2006convexity} commonly work well in low noise ratio problems, in contrast to the high-noise ratio nature of our ultrasound data. 

Active learning methods~\cite{hoi2008semi, lorente2014active, kuestner2018machine} coupled with uncertainty sampling strategy~\cite{lewis1994heterogeneous, lewis1994sequential} can detect hard informative samples and promote human observer to label only those samples.
It does not only reduce the labelling cost of non-informative samples but also guide of human observers to rectify labelling errors. 
While those methods are effective at reducing the variability in labelling, they do not model the variability.

From another point of view, observer variability can be viewed as a type of uncertainty, namely the \emph{aleatoric} uncertainty, where the aim is to capture the noise inherent in the observations through a Bayesian approach, \ie, learning the variability distribution as an output of a model~\cite{der2009aleatory, kendall2017uncertainties}.
In a clinical labelling context, this assumes the observer's \emph{gold standard} opinion, regarding a single sample subject, follows a distribution over the available options, and the model learns to estimate this distribution of opinions.

Another type of uncertainty is called the \emph{epistemic} uncertainty.
This type of uncertainty is introduced by the learning model, which can be explained away given enough data; thus it is also known as \emph{model uncertainty}.
Several Bayesian inference approaches and more recent Bayesian neural networks (BNN)~\cite{denker1991transforming, mackay1992practical, perez2007misclassified, neal2012bayesian} are designed to address the uncertainty in the induced classifier by imposing a prior distribution over model parameters.
Nevertheless, the Bayesian methods usually have a low convergence rate, which may not be suitable for solving large-scale problems.

\begin{table*}[!htbp]
    \centering
        
    \caption{The composition of the subset of data corresponding to the $\mathbf{A_1}$ and $\mathbf{A_2}$ label sets, in terms of number of studies from each type of the 14 standard echocardiography views (A\#C: apical \#-chamber view, PLAX: parasternal long axis view, RVIF: right ventrical inflow view, S\#C: subcostal \#-chamber view, IVC: subcostal inferior vena cava view, PSAX-A: parasternal short axis view at aortic valve, PSAX-M: PSAX view at mitral annulus valve level, PSAX-PM: PSAX view at mitral valve papillary muscle level, PSAX-APEX: PSAX view at apex level, and SUPRA: suprasternal view).}
    \label{tb:data_dist}
     \resizebox{\textwidth}{!}{%
    \begin{tabular}{|c|c|c|c|c|c|c|c|c|c|c|c|c|c|c|c|}
\hline																												
View	&	A2C	&	A3C	&	A4C	&	A5C	&	PLAX	&	RVIF	&	S4C	&	S5C	&	IVC	&	PSAX-A	&	PSAX-M	&	PSAX-PM	&	PSAX-APEX	&	SUPRA	\\
\hline																													
Training set	&	335	&	283	&	359	&	128	&	390	&	131	&	172	&	29	&	218	&	401	&	388	&	187	&	63	&	46	\\
\hline																													
Validation set	&	126	&	101	&	105	&	42	&	131	&	49	&	77	&	5	&	67	&	135	&	137	&	60	&	19	&	13	\\
\hline																													
Test set	&	106	&	95	&	93	&	44	&	137	&	29	&	49	&	15	&	56	&	108	&	147	&	73	&	13	&	11	\\
\hline				
    \end{tabular} %
    
    }

\end{table*}

In this paper, we propose a new method for modelling label uncertainty in data, resulting from intra-observer variability in labelling. We use 2D echo quality assessment as a test-case to demonstrate the efficacy of this modelling approach, where there is large intra-observer variability present. Furthermore, we aim to solve a regression problem where only categorical expert labels are provided.
Our proposed method, namely the Cumulative Density Function Probability (CDF-Prob) method, addresses the observer variability as aleatoric uncertainty, which models experts' opinions as Laplace or Gaussian distributions over the regression space~\cite{nix1994estimating}.
Our approach is easy to implement, and only requires replacing the \emph{Softmax} layer of a classification model. Hence, it can be applied to a wide range of clinical problems, where labels are categorical (\ie, degrees of pathology severity), and subject to large intra/inter-observer variability in gold standard labels.

Our contributions are:
\begin{itemize}
    \item we conduct an extensive study of the intra-observer variability with an archived echo image dataset labelled by an expert cardiologist, and we show with evidence that repeat labelling of the same set of data (and training with \emph{soft targets}) can be much more helpful than acquiring and labelling a larger volume of data which could be less economical;
    \item with empirical evidence, we show that the intra-observer variability is a type of aleatoric uncertainty, and we propose to model the variability in a regression setting with the use of an aleatoric uncertainty modelling approach, namely the PDF-Prob~\cite{nix1994estimating} method.
    We demonstrate that this modelling approach improves the performance compared to the conventional regression method;
    \item we also propose CDF-Prob, an extension of the PDF-Prob method that uses categorical labels, which further improves the classification performance.
\end{itemize}

\begin{figure*}[!htbp]
    \centering
    \resizebox{0.9\textwidth}{!}{%
    \begin{tabular}{cc}
        \includegraphics[width=0.6\textwidth]{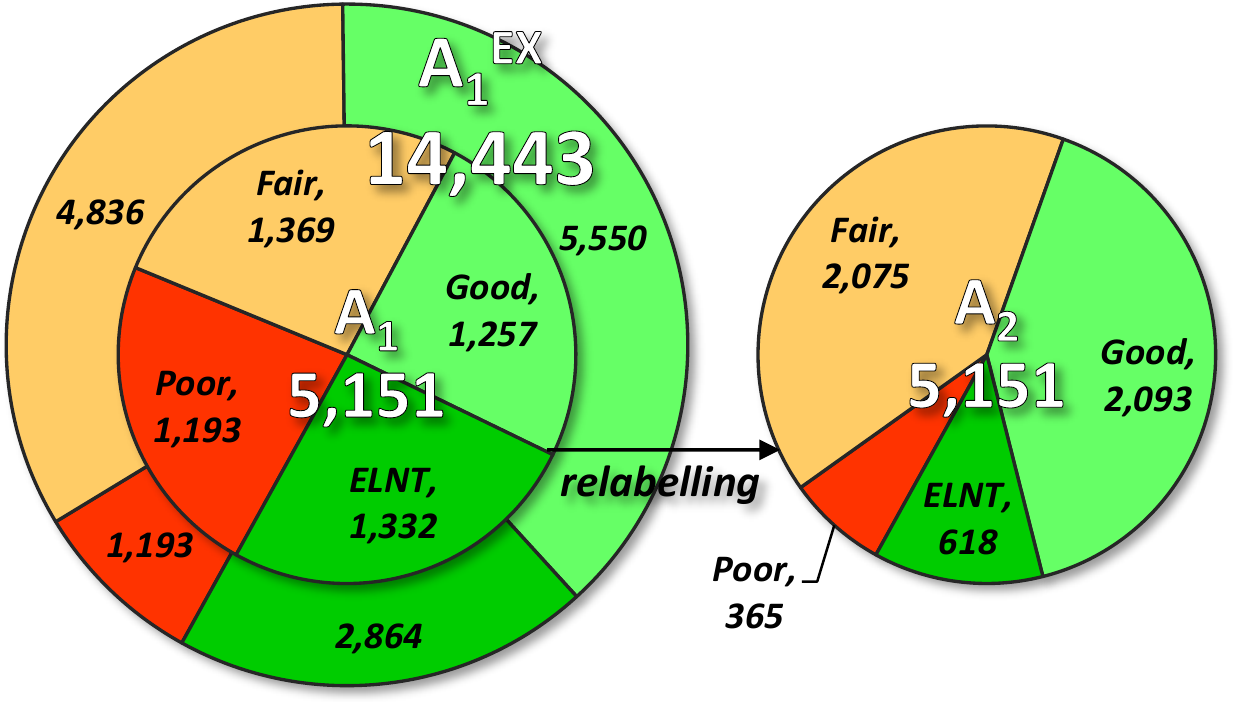}
        &  
        \includegraphics[width=0.4\textwidth]{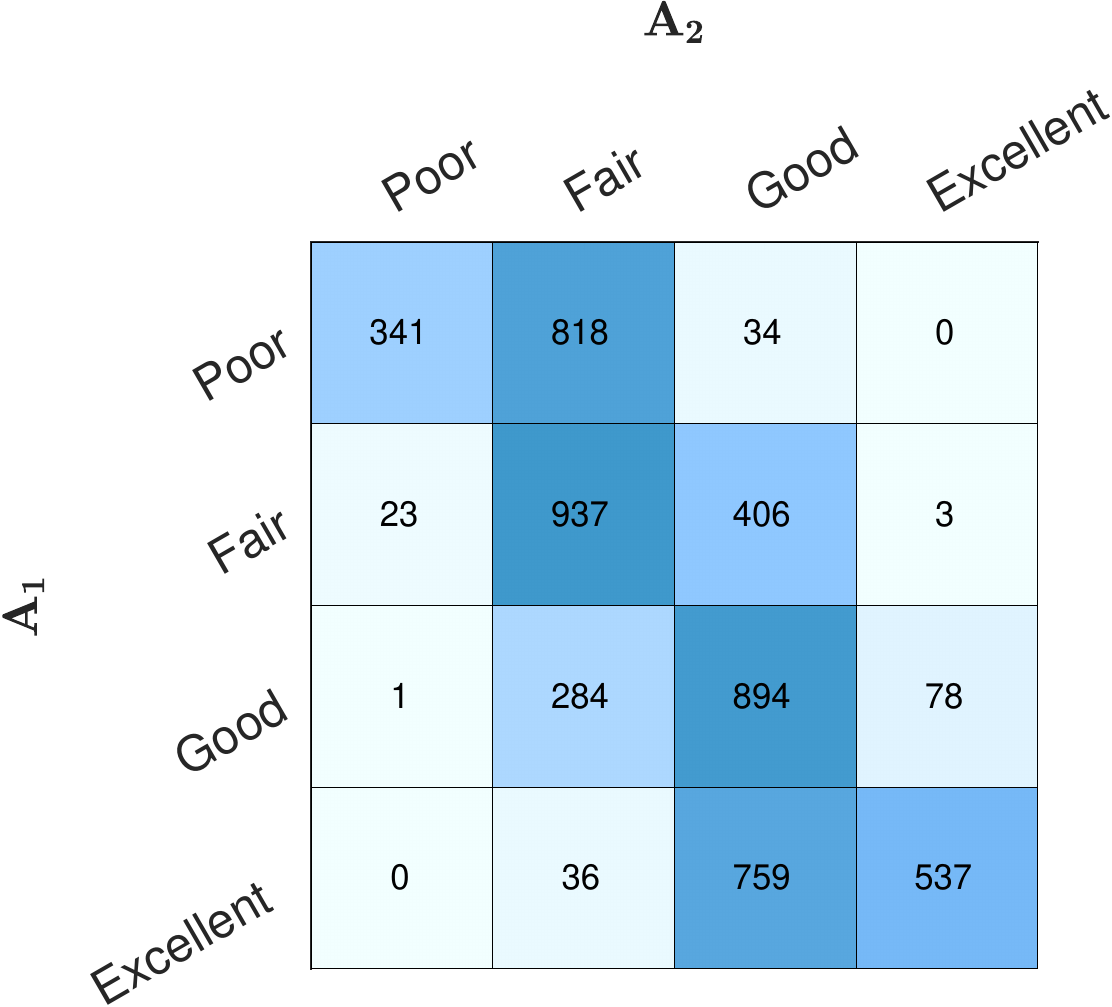}
        \\
        (a) & (b) \\
    \end{tabular}%
    }
    \caption{The discrepancy between ground truth label sets $\mathbf{A_1}$ and $\mathbf{A_2}$, showing the label set creation process in (a) and the intra-observer variability confusion matrix in (b).
    \emph{ELNT} abbreviates the \emph{Excellent} class. 
    }
    \label{fig:confusion_matrix}
\end{figure*}

\section{Problem Definition}

\subsection{Motivation}

This work is a part of a project aiming at providing real-time expert-level feedback for novice US machine operators to enhance their skills based on the images they acquired.
Our preliminary studies show that it is helpful to provide a real-time and responsive numerical feedback rather than a less responsive and discrete (categorical) feedback to operators, \ie, 
an operator can learn much better if slight changes to the position of their probe results in an increase or decrease of the quality reading.
The goal of this work is to solve this regression problem.

In an earlier attempt to acquire labels for this echo cine series quality estimation problem, multiple expert cardiologists (experience level between PGY-3 and 30 years of professional practice) were given the same task of labelling quality with a number between 0\% (poor) and 100\% (excellent) based on the standard echo image quality assessing system, \ie, the Gaudet chart system (see App.~\ref{app:gaudet})~\cite{gaudet2016focused}, quantified with an interval of 5\%. 
However, this detailed label schema resulted in elevated inter-observer variability among the cardiologists and was labour intensive.
The diversity in experts' opinions follows the fuzziness in human reasoning~\cite{zadeh1997toward}.

To address these issues, we provided another labelling schema to ask a senior cardiologist with 30 years clinical experience to label the echo cine series in one of four categories, aiming at eliminating the inter-observer variability,
The four categories are \emph{Excellent} (75\%-100\%), \emph{Good} (50\%-74\%), \emph{Fair} (25\%-49\%), or \emph{Poor} ($<$25\%), where the numerical percentage is in reference to the top achievable image quality (equivalent to the maximum score achieved in the Gaudet point system).
The determination of the four-category scheme and the labelling guideline are explained in App.~\ref{app:4class}.
The coarse labelling approach reduces the time and effort spent by the senior cardiologist to pursue fine-grain labels. 
However, it results in a regression problem with categorical labels.

\subsection{Dataset}
\label{sec:dataset}
A total of 14,443 echo studies from 3,157 unique patients (including 14 standard echocardiography views~\cite{cardiac_views}), were randomly collected from Vancouver General Hospital Picture Archiving and Communication System (PACS) for this project, under the approvals from the institutional Medical Research Ethics Board and the Information Privacy Office.
Each study is an echo cine series that contains a variable number of frames from 10 to 623 frames, where the mean number of frames is 48.
In addition, the collected studies were generated from seven different ultrasound machine models: Philips iE33, GE Vivid 7, Vivid i, Vivid E9, Sequoia, and Sonosite.
Therefore, the resolution, size of the ultrasound visual area, the probe specification, and imaging settings vary across the machine models.

The entire 14,443 studies were labelled first by the senior cardiologist for the quality category and the echocardiography view class, which took two weeks to complete. During this process, the cardiologist was viewing the original image data extracted from the archived DICOM files.
We denote the quality label set of the entire dataset as $\mathbf{A_1^{EX}}$.
It can be noted from Fig.~\ref{fig:confusion_matrix}-(a), $\mathbf{A_1^{EX}}$ has imbalanced quality distribution, hence we first applied class-wise weighting and stratified batch methods to re-balance the training.
Nevertheless, the top-1 classification accuracy of the plain regression model (specified in Sec.~\ref{sec:regression}) was not improved, leading to the investigation of the labels.
Two months later, there were 5,151 studies randomly selected from $\mathbf{A_1^{EX}}$ (the selection is at the patient level with the aim to having roughly similar amount of studies in each quality category) and relabelled with quality only by the same cardiologist to determine the level of intra-observer variability. 
The relabelling attempt took four days to complete.
The two quality label sets are distinguished as $\mathbf{A_1}$, where $\mathbf{A_1} \subset \mathbf{A_1^{EX}}$, and the second label set as $\mathbf{A_2}$, where $|\mathbf{A_1}| = |\mathbf{A_2}| = 5,151$.
The relabelling process is also depicted in Fig.~\ref{fig:confusion_matrix}-(a).
The 5,151 studies were randomly split into training (60\%), validation (20\%), and test (20\%) sets with unique patients, resulting in 3,118, 1,062, and 971 mutually exclusive studies distributed in respective sets (see Table~\ref{tb:data_dist} for the distribution of studies in respective view classes).
The remaining 9,292 studies were used as additional training data depending on the training scenarios described in Sec.~\ref{sec:scenarios}.

\begin{figure*}[!htbp]
    \centering
    \includegraphics[width=0.75\textwidth]{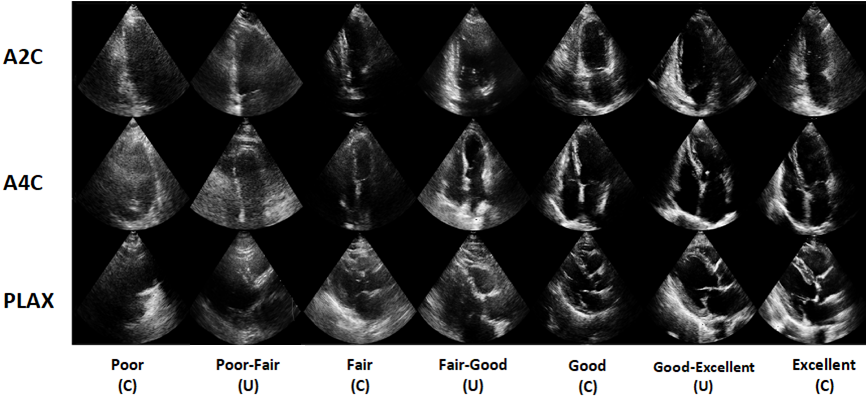}
    \caption{Example image frames of study cine series ordered from \emph{Poor} to \emph{Excellent} quality classes;
    (C) indicates certain and (U) indicates uncertain studies.
    In general, adjacent certain and uncertain study samples are visually indistinguishable.
    However, it is easy to see a general blurry to clear transition of cardiac structure from left to right by a non-expert viewer.}
    \label{fig:quality_example}
\end{figure*}

$\mathbf{A_1}$ and $\mathbf{A_2}$ have 47.4\% inconsistent labelling, where 45.9\% are one category mismatch. 
We infer the confusion also comes from the fuzziness in the human reasoning towards the concept of quality, even the categories are defined with explicit boundary values.
Beyond that, the relative difference between the displayed consequent studies may also cause the cardiologist to label based on the relativity.
Furthermore, the quality difference between the frames in a cine series (\ie, the hand of a sonographer may have moved when he or she recorded the study, causing inconsistent quality over a cine series).
In Fig.~\ref{fig:confusion_matrix}-(b) and (c), we illustrate the discrepancy between $\mathbf{A_1}$ and $\mathbf{A_2}$ on the entire 5,151 studies.
We further define the concept of a {\bf certain study} if the labels from $\mathbf{A_1}$ and $\mathbf{A_2}$ are identical for a study, and an {\bf uncertain study} if not.
In Fig.~\ref{fig:quality_example}, we show an example of study cine series frames from three of the echo view classes.

\section{Methodology}

For clarity, a list of the symbols used in this paper are summarized in Table~\ref{tb:symbols}.

\begin{table}[!htbp]
    \centering
    \caption{Table of Symbols.}
    \resizebox{0.9\columnwidth}{!}{%
    \begin{tabular}{ccr}
    \hline \hline
    Symbol & Set notation & Explanation \\
    \hline \hline
    - & $\mathbf{W}$ & model parameter \\
    \hline
    - & $\mathcal{D}$ & dataset \\
    \hline
    $x$ & $\mathbf{X}$ & image \\
    \hline
    $y$ & $\mathbf{Y}$ & regression label/expectation of $e$\\
    \hline 
    $\sigma$ & - & standard deviation of $e$\\
    \hline
    $e$ & $\mathcal{E}$ & expert opinion distribution $\mathcal{N}(y, \sigma^2)$\\
    \hline
    $f$ & - & model producing estimated $\hat y$\\
    \hline
    $g$ & - & model producing estimated $\hat \sigma$\\
    \hline
    c & $\mathcal{C}$ & categorical label \\
    \hline
    $l$ & $\mathcal{L}$ & $c$'s numerical lower bound \\
    \hline
    $u$ & $\mathcal{U}$ & $c$'s numerical upper bound \\
    \hline
    $a_1$ & $\mathbf{A_1}$ & 1$^\text{st}$ trial categorical label \\
    \hline
    $a_2$ & $\mathbf{A_2}$ & 2$^\text{nd}$ trial categorical label \\
    \hline \hline
    \end{tabular}%
    }
    \label{tb:symbols}
\end{table}

\subsection{Uncertainty Modelling}
\label{sec:related_works}

As mentioned, uncertainty can be categorized into epistemic and aleatoric uncertainties~\cite{kendall2017uncertainties}.
Epistemic uncertainty is the property of a learning model induced by 
limited data;
therefore it can be explained away by learning from a larger set.
With limited data availability, one can use Bayesian neural networks (BNN)~\cite{denker1991transforming, mackay1992practical, perez2007misclassified, neal2012bayesian} to model epistemic uncertainty.
BNNs utilize Bayesian inference to model the posterior distribution of the neural network parameters given the data.
Let a dataset be noted as $\mathcal{D} = \{ \mathbf{X}, \mathbf{Y} \}$, where $\mathbf{X} = \{ \mathbf{x}_i\}_{i=1}^{|\mathcal{D}|}$ and $\mathbf{Y} = \{ y_i\}_{i=1}^{|\mathcal{D}|}$ denote the sets of $|\mathcal{D}|$ observed samples and corresponding labels, and the posterior of model parameters $\mathbf{W}$ is formulated as $p(\mathbf{W} | \mathcal{D} ) = \frac{p(\mathbf{Y} | \mathbf{X}, \mathbf{W})p(\mathbf{W}|\mathbf{X}) }{p(\mathbf{Y} | \mathbf{X})}$.
Since the marginal distribution $p(\mathbf{Y} | \mathbf{X})$ is intractable, various  methods approximate $p(\mathbf{W} | \mathcal{D} )$ with a simple distribution $q_{\theta}(\mathbf{W})$, parameterized by $\theta$~\cite{graves2011practical, gal2015bayesian, blundell2015weight, gal2016dropout}.
Particularly, Monte-Carlo Dropout (MC-Dropout) is one of the commonly used Bayesian inference approximation method~\cite{gal2016dropout}, which has been used in medical imaging to derive robust solutions for disease detection, lesion/organ segmentation~\cite{leibig2017leveraging, nair2018exploring, roy2019bayesian}.
During the training phase of MC-Dropout, model parameters sampled from $q_{\theta}(\mathbf{W})$ are used to minimize a task-related loss function, where the introduction of Dropout minimizes Kullback-Leibler (KL) divergence between the approximate posterior $q_{\theta}(\mathbf{W})$ and true posterior $p(\mathbf{W} | \mathcal{D})$~\cite{srivastava2014dropout}.
During inference phase, the model parameters are also sampled multiple times from $q_{\theta}(\mathbf{W})$ to compute an uncertainty interval in the network prediction.

The aleatoric uncertainty, on the contrary, describes the confusion nature of $\mathcal{D}$, which cannot be explained away even if a larger dataset is available.
In our problem, we assume the intra-observer variability as $y_i \xleftarrow[]{} e_i$, \ie, $y_i$ is a random sample drawn from an ``expert opinion'' distribution $e_i$ towards a \emph{single} study $\mathbf{x}_i$ (the subscript $i$ is dropped hereafter to note a single sample).
Nix and Weigend~\cite{nix1994estimating} formulated least-squares regression as a maximum likelihood estimation problem 
with an underlying Gaussian error model, \ie,
$e \sim \mathcal{N}\big(f(\mathbf{x}; \mathbf{W}_f), g^2(\mathbf{x}; \mathbf{W}_g)\big)$, where $f$ and $g$ denote two functions that compute the mean and standard deviation (STD) parameters respectively, and $\mathbf{W}_f \cup \mathbf{W}_g = \mathbf{W}$ denote model parameters.
As an example, Nix and Weigend~\cite{nix1994estimating} used a Gaussian probability density function to approximate the likelihood of the expert opinion distribution:
\begin{equation}
    p(e | \mathbf{x}, \mathbf{W}) = \frac{1}{\sqrt{2 \pi g^2( \mathbf{x} )}} \exp(\frac{-||y - f( \mathbf{x} ) ||^2}{2g^2( \mathbf{x} )}),
\label{eq:pdf}
\end{equation}
and the training objective minimizes the empirical negative log-likelihood $\mathbb{E}_{\mathbf{x} \in \mathbf{X}}[- \ln p(d | \mathbf{x}, \mathbf{W})]$:
\begin{equation}
 \ell(\mathbf{W}, \mathcal{D}) = - \frac{1}{|\mathcal{D}|} \sum_{i=1}^{|\mathcal{D}|} \frac{1}{2}\big(\frac{||y_i - f(\mathbf{x}_i)||^2}{g^2(\mathbf{x}_i)} + \ln g^2(\mathbf{x}_i)\big).
 \label{eq:aleatoric_regression}
\end{equation}
An issue raised with Eq.~(\ref{eq:pdf}) is that the computed values are probability densities on the observed points, and the density values can go beyond 1.
In our specific problem, the label is provided in a categorical format, \eg, $c$ = \emph{Fair} defines an interval $y \in (0.25, 0.5]$, which enables us to compute the exact probability value of the class interval instead of approximating the density value from the class center (at the point $y = 0.375$).

\begin{figure*}[!htbp]
    \centering
    \resizebox{1.0\textwidth}{!}{%
    \begin{tabular}{ccc}
        \includegraphics[width=0.352\textwidth]{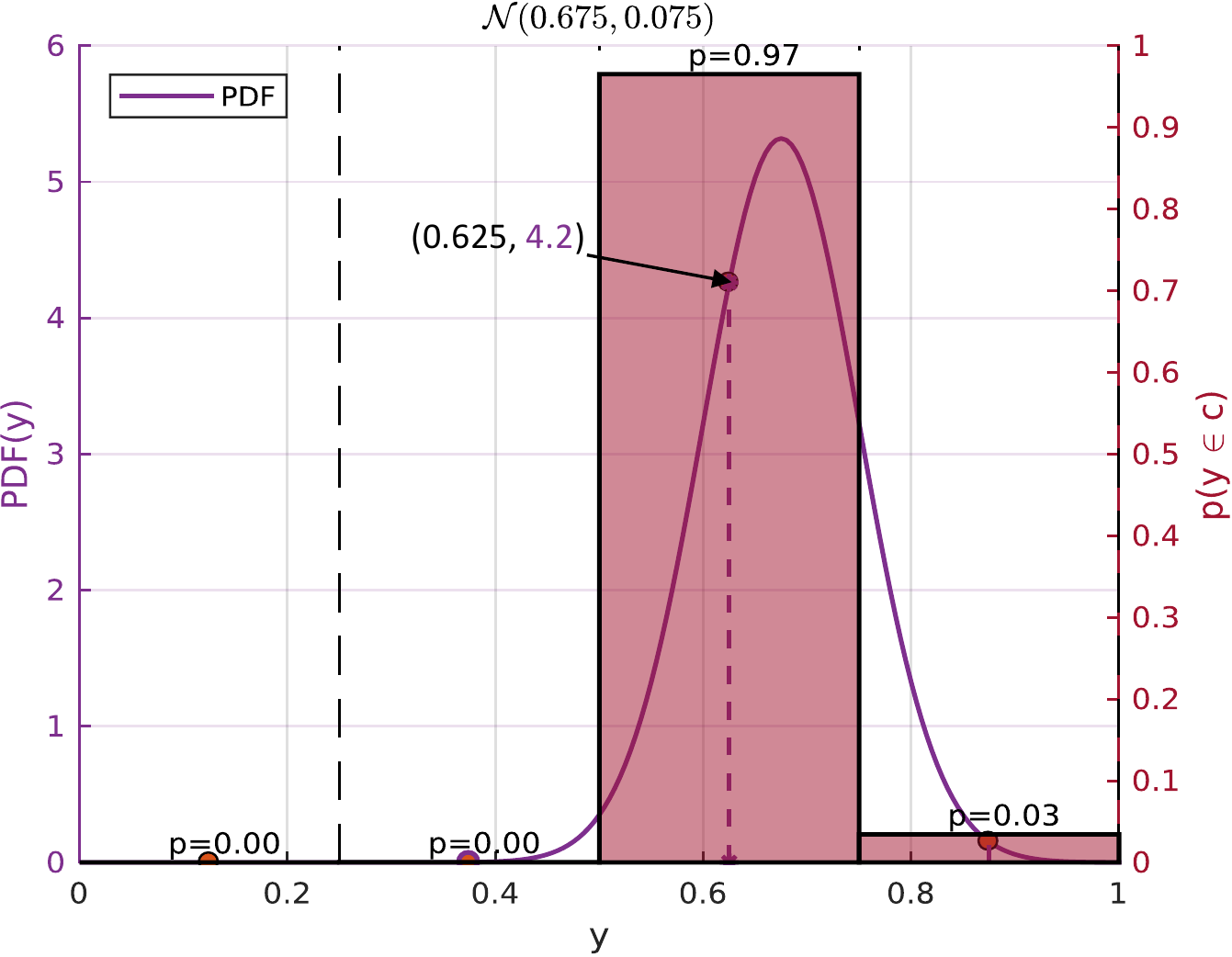} &  
        \includegraphics[width=0.33\textwidth]{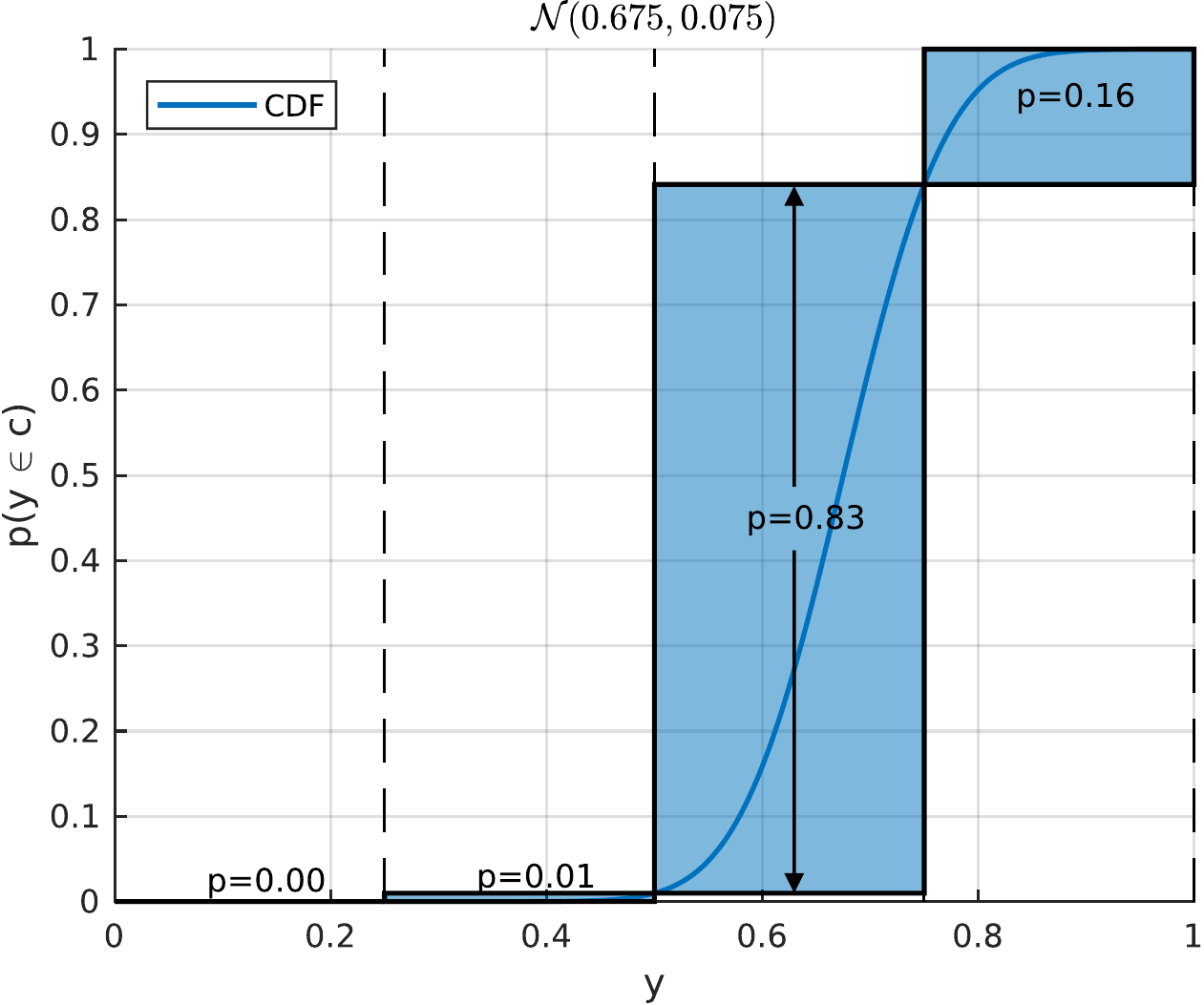} &
        \includegraphics[width=0.33\textwidth]{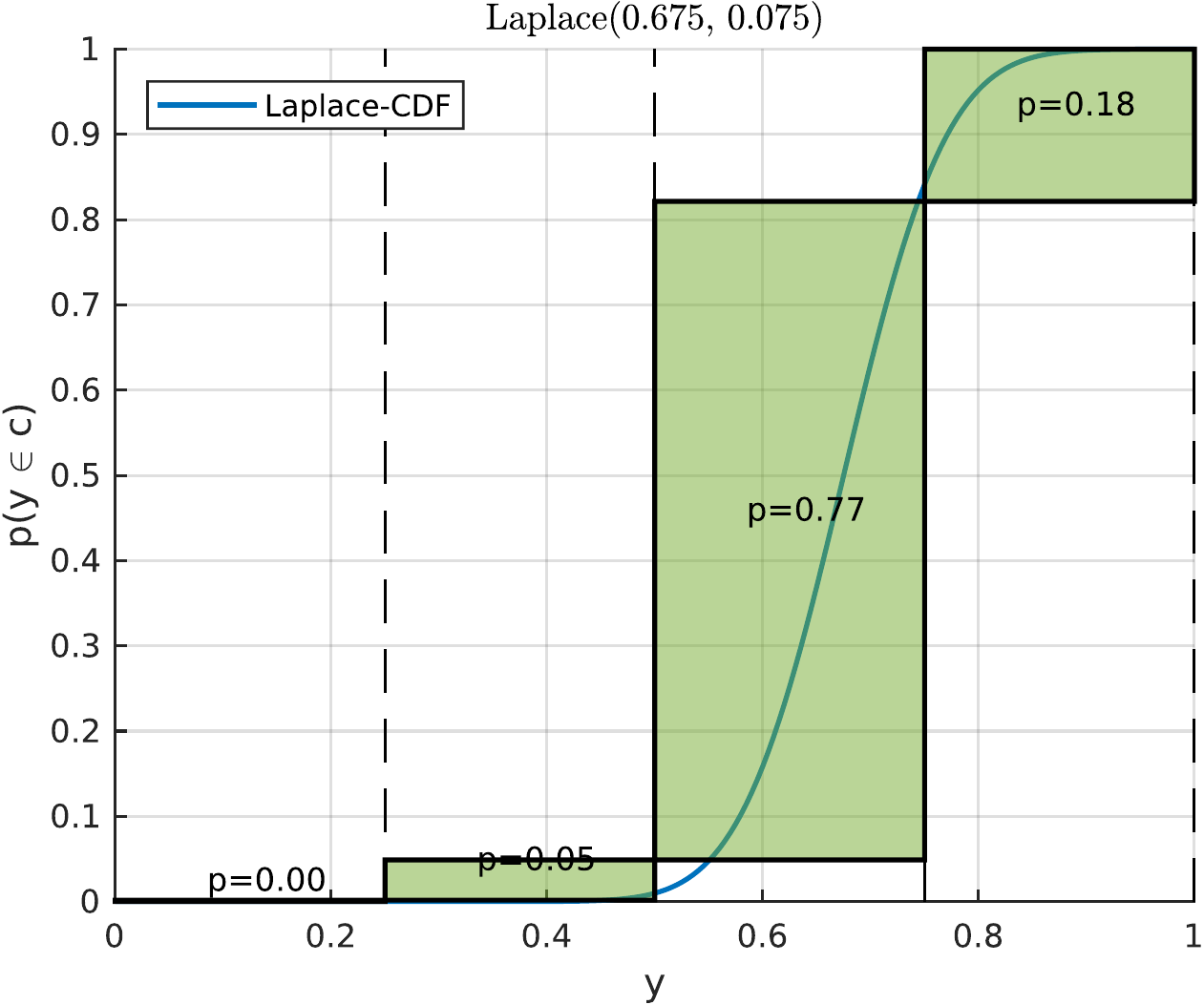} \\
        (a) & (b) & (c)\\
    \end{tabular}%
    }
    \caption{A comparison of class-wise likelihood values computed by (a) the density values (approximated by Eq.~(\ref{eq:pdf}) with the use of the center point of each class, represented by the purple dot lines), and (b) the true probability values (computed by Eq.~(\ref{eq:true_distribution})), and (c) the Laplace distribution computed probability values (assuming $g(\mathbf{x})$ yields the \emph{scale} parameter), all of which are normalized to ensure unit sum of class probabilities, for an example case that assuming $c = \textit{Good}$. 
    In (a), the most likely observed class $c = \textit{Good}$ has density 4.2, and the normalized probability is 0.97. 
    Given the same parameters, the Laplace density computed probabilities in (c) are much more softly distributed in respective classes than in (b).
    } 
    \label{fig:pdf_cdf_comparison}
\end{figure*}

\subsection{Multi-label Problem}

The label discrepancy in $\mathbf{A_1}$ and $\mathbf{A_2}$ can be considered as a type of multi-label problem.
For odd numbers of sets of labels, one can use \emph{majority voting} to determine the label for the uncertain studies.
Nevertheless, for an uncertain study with two sets of labels, it is possible to handle the discrepant labels by soft targets~\cite{hinton2015distilling} method.
The soft targets methods~\cite{hinton2015distilling, romero2014fitnets} are used as regularization techniques for training a small model with the predictions inferred from a high complexity model or an ensemble of them (trained from the same dataset), for the purposes of capturing the knowledge of the high complexity models with fewer model parameters (in our case, the knowledge is the cardiologist's opinion).
The soft targets method is therefore explored in this work to address the intra-observer variability problem (In this paper, we refer to the soft targets as the Average Ground Truth method, see Sec.~\ref{sec:loss}).

It should be pointed out that the soft targets method resembles the label distribution learning (LDL) methods~\cite{geng2016label, gao2017deep}, where the essence of LDL methods is to to match a likelihood distribution estimation (by a model) to the ground truth label distribution by using KL divergence.
On the other hand, the soft targets method matches the same by cross entropy.
The labels are fixed so that only the likelihood distribution is to be learned; therefore, the objective of KLD and cross-entropy is equivalent in this case, \ie, their learning objective is with a constant offset difference produced by the entropy of the label distribution.

\subsection{Incorporating the categorical label}

To be consistent with the notations defined in Sec.~\ref{sec:related_works}, our dataset is defined as $\mathcal{D} = \{\mathbf{X}, \mathbf{A_1}, \mathbf{A_2}\}$, where $\mathbf{X}$ is the set of images as previously defined, and $\mathbf{A}_1 = \{{a_1}_i\}_{i=1}^{|\mathcal{D}|}$ and $\mathbf{A}_2 = \{{a_2}_i\}_{i=1}^{|\mathcal{D}|}$, are the two sets of categorical labels annotated by the cardiologist, and ${a_1}_i, {a_2}_i \in \mathcal{C} = \{ c : \textit{Poor}, \textit{Fair}, \textit{Good}, \textit{Excellent} \}$.
Our goal is to determine a set of expert opinion distributions $\mathcal{E} = \{e_i\}_{i=1}^{|\mathcal{D}|}$, where $e_i \sim \mathcal{N}(y_i, \sigma^2_i)$
represents the error of aleatoric uncertainty modelled as Gaussian noise with expectation value being the true label $y_i \in (0, 1]$ and certain variation $\sigma$ encodes the uncertainty of the expert opinions (\ie, the different labels obtained in different trials) that are both bounded to a standard normal distribution $\mathcal{N}$.

We model $y$ and $\sigma$ with two deep logistic regression models $\hat y = f(\mathbf{x}; \mathbf{W}_f)$ and $\hat \sigma = g(\mathbf{x}; \mathbf{W}_g)$ to ensure a proper value range $(0, 1]$.

\begin{figure}
    \centering
    \includegraphics[width=0.9\columnwidth]{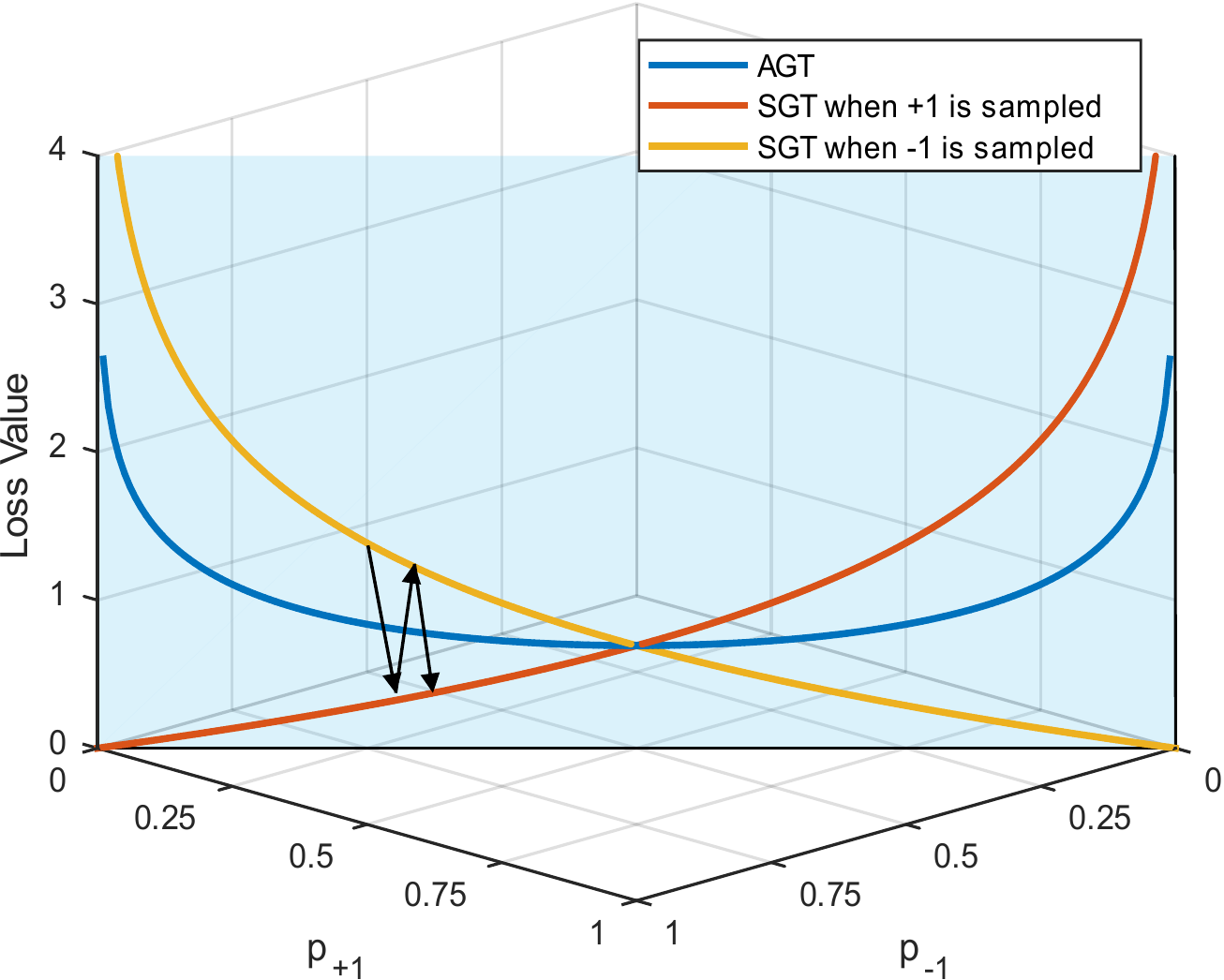}
    \caption{Comparison of AGT and SGT generated loss signals with the use of a two-class example (\{+1, -1\}), shown as a function of probabilities of both classes, for an uncertain case $p_{+1} = p_{-1} = 0.5$.
    The loss signal of SGT method can be unstable due to the sampling process (\ie, jumping between the red and yellow lines), while AGT loss signal follows a stable averaging of the two SGTs. 
    }
    \label{fig:loss_comparison}
\end{figure}

\subsection{Likelihood Probability}

Given that both $\mathbf{A_1}$ and $\mathbf{A_2}$ are categorical and we know the class lower and upper bounds in the regression space $ \mathcal{L} = \{ l_c : c \in \mathcal{C} \} = \{0, 0.25, 0.5, 0.75\}$ and $\mathcal{U} = \{ u_c : l_c + 0.25\}_{c \in \mathcal{C}}$, the cumulative density function (CDF) of a number $z$ with respect to (w.r.t.) $f(\mathbf{x})$ and $g(\mathbf{x})$ can be written as:
\begin{align}
\label{eq:cdf}
    F(z) & = \frac{1}{2} \big( 1 + \text{erf}(\frac{z - f(\mathbf{x})}{g(\mathbf{x}) \sqrt{2}})\big), \text{\quad and} \\ \nonumber
    \text{erf}(z) & = \frac{2}{\sqrt{\pi}} \int_0^z \exp(-t^2) dt,
\end{align}
where $\text{erf}(z)$ is the error function, for computing the probability of $ y \in (-\infty, z]$, for $z \in \mathbb{R}$.
Thus, we can express $p( c | \mathbf{x}, \mathbf{W}) = p(\hat y \in (l_c, u_c]|\mathbf{x}, \mathbf{W})$ (short noted as $\hat p^*_c$) as the likelihood probability of $\hat y$ falling in the category $c$:
\begin{align}
\label{eq:cdf-prob} 
    \hat p^*_c & = F(u_c) - F(l_c) \\ \nonumber
    & = \frac{1}{2}\big( \text{erf}(\frac{u_c-f(\mathbf{x})}{g(\mathbf{x})\sqrt{2}}) - \text{erf}(\frac{l_c-f(\mathbf{x})}{g(\mathbf{x})\sqrt{2}}) \big) .
\end{align}
Since the definition of the problem forbid observations of samples with quality below \emph{Poor} or above \emph{Excellent}, but expert opinion $e$ is defined over $\mathbb{R}$, we normalize the likelihood probabilities to ensure a unit sum:
\begin{equation}
    \hat p_c = \frac{\hat p^*_c}{\sum_{c \in C} \hat p^*_c},
\label{eq:true_distribution}
\end{equation}
which fits the regression problem within a classification framework.

\begin{figure*}[!htbp]
    \centering
    \includegraphics[width=1.0\textwidth]{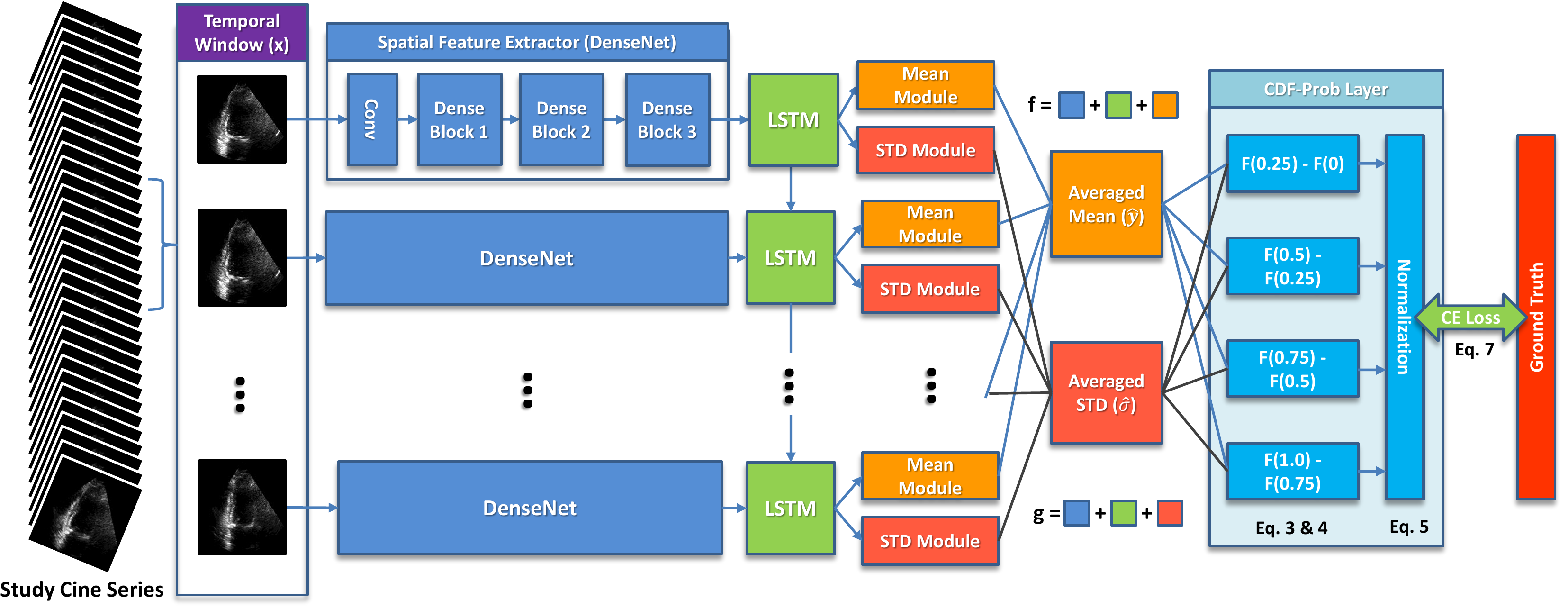}
    \caption{Schematic of the network architecture for echo quality assessment.
    In this design, the two deep logistic regression models $f$ and $g$ share the same DenseNet + LSTM feature model.
    The proposed CDF-Prob layer is detailed to illustrate the internal computation.}
    \label{fig:architecture}
\end{figure*}

\begin{figure}[!htbp]
    \centering
    \includegraphics[width=0.65\linewidth]{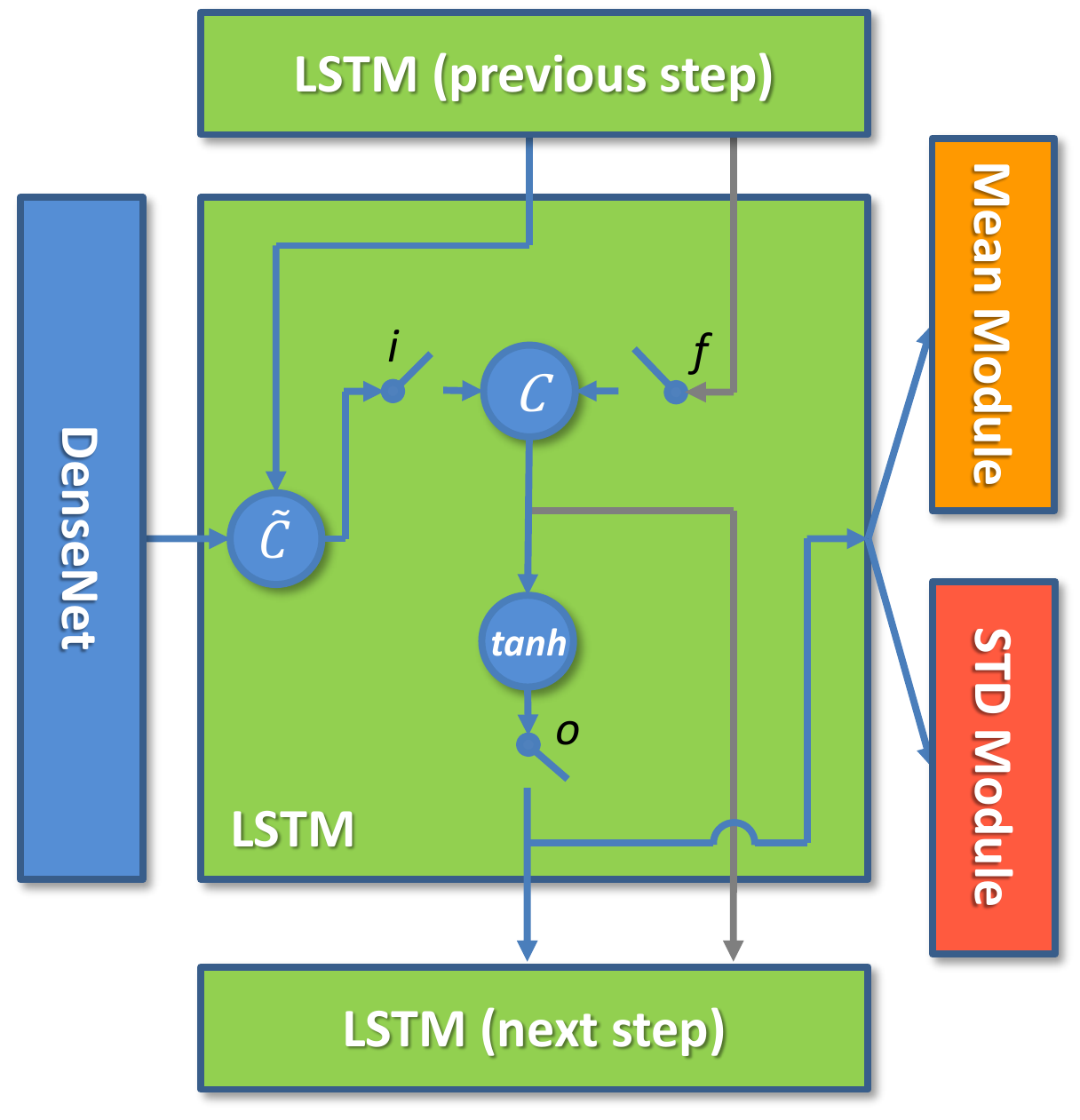}
    \caption{Schematic of the internal structure of an LSTM module, illustrated by an intermediate step. $i, f, o$ are the input, forget, and output gates. 
    The features from DenseNet and LSTM's the last step are used to compute a new candidate state $\tilde c$ jointly. 
    The previous step's cell memory (going through the gray inflow arrow) and $\tilde c$ are weighted by respective gates $f$ and $i$ and then jointly composite the new memory state $c$. Finally, the activated memory state is weighted by $o$ before being visible to the LSTM's next step and the dependent layers (\ie, the Mean and STD Modules).}
    \label{fig:lstm}
\end{figure}

In Fig.~\ref{fig:pdf_cdf_comparison}-(a) and (b), we show an example that Eq.~(\ref{eq:pdf}) results in a roughly 4.2 density value to represent the likelihood of the $\mathit{Good}$ class in (a) and the true probability 0.83 in (b) from the same Gaussian, where the respective loss values, $-\log(4.2) \approx -0.63$ and $-\log(0.83) \approx 0.08$, are distinctive, which would inevitably have different training convergence properties.
In Sec.~\ref{sec:experiments}, we provide empirical evidence that maximizing the log-likelihood of the true probability distribution has an advantage over the density distribution in terms of improved performance with the use of the same model.

In addition, this framework can be integrated with the Laplace distribution, by defining  $F_{\mathrm{Lap}}(z)$ 
as
\begin{multline}
F_{\mathrm{Lap}}(z) = \\ \frac{1}{2}\bigg(1 + \mathrm{sgn}\big(z - f(\mathbf{x})\big) 
\big(1 - \exp(-\frac{|z - f(\mathbf{x})|}{g(\mathbf{x})}) \big)\bigg).
\end{multline}
The main difference lies in that the Laplace density implements an absolute difference around mean while the Gaussian density implements a squared difference. 
The Laplace distribution computed probabilities are shown in Fig.~\ref{fig:pdf_cdf_comparison}-(c).

\subsection{Loss Function}
\label{sec:loss}

The model is trained to maximize the likelihood over $\mathbf{A_1}$ or $\mathbf{A_2}$, \ie, $\mathrm{argmax}_\mathbf{W}\; \mathbb{E}_{\mathbf{x} \in \mathbf{X}} [p(a_1 \lor a_2| \mathbf{x}, \mathbf{W})]$, through the cross-entropy minimization:
\begin{equation}
\ell(\mathbf{W}, \mathcal{D}) = - \frac{1}{|\mathcal{D}|} \sum_{i=1}^{|\mathcal{D}|} \left( 
{\begin{aligned}
    & \;\;\;\; \lambda_1\log p( {a_1}_i | \mathbf{x}_i, \mathbf{W}) \\
    & + \lambda_2\log p( {a_2}_i | \mathbf{x}_i, \mathbf{W})
\end{aligned}}\right),
\label{eq:cross_entropy_loss}
\end{equation}
where the scalars $\lambda_1$ and $\lambda_2$ are the weighting assignments for the observed classes, respectively.
In this paper, we refer to the soft targets method (\ie, $\lambda_1=\lambda_2=\frac{1}{2})$ as the  Average Ground Truth (AGT) method. This is an averaging approach in the sense that the one-hot (\ie, a binary vector with only one element being 1) representations of $a_1$ and $a_2$ are averaged, \ie, $ \frac{1}{2} ([\mathbb{1}(a_1=c)]^{\mathrm{T}}_{c\in\mathcal{C}} + [\mathbb{1}(a_2=c)]^{\mathrm{T}}_{c\in\mathcal{C}})$), where $\mathbb{1}(.)$ is the indicator function.
Another way to utilize Eq.~(\ref{eq:cross_entropy_loss}) is to feed the label stochastically, \ie, either $\lambda_1 = 1$ or $\lambda_2 = 1$, while the other is set to zero.
This acts as a standard cross-entropy loss, referred to here as Stochastic Ground Truth (SGT) method. 

In comparison, the AGT method shows an improved stability on the loss composition, while the SGT method has a better exploratory ability.
A toy example illustrating the difference between the AGT and SGT generated loss signals is depicted in Fig.~\ref{fig:loss_comparison} with the use of a two-class example (\{+1, -1\}), the details are explained in the App.~\ref{app:toy}.

\section{Experiments}
\label{sec:experiments}

\subsection{Deep Learning Architecture} 
\label{sec:network}

A single network is used to model the echo quality estimation task for all 14 views. 
The input to our network is a 10-consecutive-frame temporal window of a study cine series (\ie, $\mathbf{x}$, 10 is determined by the minimum number of the frames), where each frame is a $120 \times 120$ gray-scale image.
A DenseNet~\cite{huang2017densely} model, severed as feature extractor, is used to extract per-frame image feature. 
The DenseNet model has three dense blocks, each followed by a Dropout~\cite{srivastava2014dropout} layer and an average-pooling layer. 
The convolution layer before the first dense block has 12 output channels, where the output of each dense block increases the number of output channels from the previous block by 6, reaching to 30 output channels after the third dense block.
Throughout the DenseNet model, no bottle-neck convolution layer is used.
The convolution layers share the same specification, \ie, $3 \times 3$ filter size with stride 1, no padding, batch-normalized~\cite{ioffe2015batch}, and use ReLU~\cite{nair2010rectified} activation function.
The DenseNet processed features from the 10 consecutive frames are passed to an LSTM~\cite{hochreiter1997long} layer with 128 \emph{tanh} units (see Fig.~\ref{fig:lstm}), whose function is to compose an embedding for the extracted spatial and temporal features, from the processed cine frames.\footnote{The symbols defined in Fig.~\ref{fig:lstm} may have been repetitively defined in the main article.}
It is then followed by two logistic regression modules, namely the Mean and STD modules, to compute the per-frame $\hat y$ and $\hat \sigma$ estimations.
The final predictions for the entire window are averaged over the frame-wise predictions.
The total number of parameters in this network is 3.5 million.
The overview of the network architecture is depicted in Fig.~\ref{fig:architecture}. 

The proposed method is a computation layer that translates the $\hat y$ and $\hat \sigma$ to a set of probabilities, which replaces the use of a \emph{Softmax} layer.
Our implementation combines Keras (for building the network) and TensorFlow (for composing the custom CDF-Prob layer and automatically computing the gradient), where the $\mathrm{erf}$ function has an off-the-shelf implementation in TensorFlow, \ie, \textrm{tf.math.erf}.
Note that the proposed DenseNet + LSTM model is only for demonstration purpose to establish a baseline, where other types of CNN and RNN models may substitute our selections and yield better performance.
Nevertheless, it should be noted that the performance variations may be subtle when the deployed number of units/parameters are similar~\cite{chung2014empirical, yin2017comparative, dezaki2018cardiac}.

\subsection{Training hyper-parameters} 
Given the differences in image settings and imaging systems, we prepared the study cine series by semi-automatically cropping the ultrasound beam and down-sizing all frame sizes to $120 \times 120$ pixels, gray-scale and real-valued images (see App.~\ref{app:semi}).
No other pre-processing techniques have been applied to the input images.
The Adam~\cite{kingma2014adam} optimizer is used to train this network end-to-end from scratch.
The initial learning rate is set to 2.5e-4, decaying by scale 0.91 every two epochs, till it decays to approximately 100 times smaller at the 100$^{\text{th}}$ epoch. For the PDF-Prob method (see Sec.~\ref{sec:evaluated_methods}), the initial learning rate is 10 times smaller as training with a larger learning rate does not converge for this method.
During the training phase, the input data are augmented by using random translation up to $10\%$ of image dimensions in pixels and random rotation up to $\pm 5$ degrees. For the test phase, the input data are un-augmented. 
Also, the final prediction of a test cine series is pooled from the predictions of five randomly sampled temporal windows, where the windows can have overlapped frames.
Finally, weight decay is set to 5e-4.
Note that the aforementioned hyper-parameters were validated on the validation set first, then applied to train a set of new model instances with the combined training and validation sets. 
The reported experiment results in this section are computed from the test set.

\subsection{Training Scenarios}
\label{sec:scenarios}

The availability of two sets of labels enable us to explore different training scenarios to investigate the influence of information availability towards training generalizability.

\subsubsection{Training with only $\mathbf{A_1}$ or $\mathbf{A_2}$ labels}
in this training scenario, only one set of labels is used to train a model, hereafter noted as $\mathrm{S_\#}$.
This training scenario yields the baseline performance in a conventional setup on a moderate sized dataset. 

\subsubsection{Training with both $\mathbf{A_1}$ and $\mathbf{A_2}$ labels}
in this scenario, models are trained with both sets of labels, hereafter noted as $\mathrm{S_{1+2}}$, which allows us to explore the effect of label stability and exploratory ability towards the performance.
The usage of a ground truth method is noted as a suffix of the scenario notation, \eg, $\mathrm{S_{1+2}}$-AGT.

\subsubsection{Training with the extended $\mathbf{A_1^{\mathrm{EX}}}$ labels}
in this scenario, noted as $\mathrm{S_1^{\mathrm{EX}}}$, we do {\bf not} use $\mathbf{A_2}$ labels. Instead, models are trained with only 
 $\mathbf{A_1^{\mathrm{EX}}}$ labels to investigate two questions: 1) can the uncertainty in labels be mitigated by using larger training data?; 2) can the methods handle the intra-observer variability without explicitly using the one-to-many mapping information?

\subsection{Test Criteria}
\label{sec:exp_test_criteria}

The main goal of the experiments is to model the intra-observer variability, where ideally the estimated distribution $\hat e$ should be measured against $e$, but $e$ is unknown.
The substitute objective is to measure classification accuracy of the prediction against the observed categorical labels on the test set: 
\begin{align}
\label{eq:accu}
    &\mathrm{acc}(\mathcal{D}_\mathrm{test}) = \\ \nonumber
    &\frac{1}{|\mathcal{D}_\mathrm{test}|} \sum_{i=1}^{|\mathcal{D}_\mathrm{test}|} \mathbb{1}\big(\hat y_i \in \big(\min(l_{{a_1}_i}, l_{{a_2}_i}), \max(u_{{a_1}_i}, u_{{a_2}_i})\big] \big),
\end{align}
where $\mathbb{1}$ is the indicator function and the individual term relaxes $\mathbb{1}(\hat c_i = {a_1}_i \lor \hat c_i = {a_2}_i)$ because we assume the continuity of the underlying distribution, \ie, ${a_1}$ = \emph{Poor} and ${a_2}$ = \emph{Good} meaning the skipped middle class \emph{Fair} is also a valid label. 
This assumption is only valid in a problem like ours, whereas in a general classification task schema, the middle class(es) between a pair of classes may not be meaningful.%

As for evaluating the regression performance, we compare the mean parameter $\hat y$ with the empirical estimation of $\bar y$ by measuring the $L_1$ distance.
For each class label $c$, we assume the corresponding regression label as $y_c = \frac{u_c + l_c}{2}$.
Due to the discrepancy in $\mathbf{A_1}$ and $\mathbf{A_2}$, we further estimate the test ground truth as $\bar y = \frac{y_{a_1} + y_{a_2}}{2}$.
Hence, the evaluation metric of the regression performance is the mean absolute error, \ie, 
\begin{equation}
    \mathrm{abs}(\mathcal{D}_{\mathrm{test}}) = \frac{1}{|\mathcal{D}_{\mathrm{test}}|}\sum_{i=1}^{|\mathcal{D}_\mathrm{test}|} |\bar y_i-\hat y_i|.
\label{eq:abs}    
\end{equation} 

In addition, we measure the \emph{expected calibration error} (ECE) and \emph{maximum calibration error} (MCE) scores.
Both measurements are indicators for identifying if a model is well calibrated (a smaller error value means better calibrated), \eg, at a level of confidence, say 0.7, statistically speaking the number of correct predictions should be close to $70\%$~\cite{guo2017calibration}.
In order to compute ECE and MCE, the predictions are quantified into $M=10$ bins of size $1/M$ based on the confidence of each prediction: $\mathrm{conf}(\mathbf{x}) = \max_{c \in \mathcal{C}} p(c | \mathbf{x}, \mathbf{W})$.
Hence, we can define $\mathcal{B}_m = \{ \{\mathbf{x}_i, {a_1}_i, {a_2}_i\} : \mathrm{conf}(\mathbf{x}_i) \in (\frac{m-1}{M},\frac{m}{M}) \}_{i=1}^{|\mathcal{D}_\mathrm{test}|}$ to note the subset of samples that fall in the $m^{\text{th}}$ bin.
The ECE is then measured as:
\begin{equation}
    \mathrm{ece}(\mathcal{D}_{\mathrm{test}}) = \sum_{m=1}^{M} \frac{|\mathcal{B}_m|}{|\mathcal{D}_\mathrm{test}|} |\mathrm{acc}(\mathcal{B}_m) - \mathrm{conf}(\mathcal{B}_m)|,
\end{equation}
where $\mathrm{acc}(.)$ refers to the same relaxed accuracy definition in Eq.~(\ref{eq:accu}).
Finally, the MCE is computed as:
\begin{equation}
    \mathrm{mce}(\mathcal{D}_{\mathrm{test}}) = \max_{m\in \{1, \ldots, M\}} |\mathrm{acc}(\mathcal{B}_m) - \mathrm{conf}(\mathcal{B}_m)|.
\end{equation}

\subsection{Evaluated Methods}
\label{sec:evaluated_methods}

We evaluate four different methods within the aforementioned training scenarios and test criteria.

\subsubsection{Regression}
\label{sec:regression}
The baseline method is the plain regression method, where the categorical label is also translated to regression label $\bar y$ for training as it is described in Sec.~\ref{sec:exp_test_criteria}.
The prediction confidence of a sample is approximated by the inverse of the distance between $\hat y$ to each class center: $p(c | \mathbf{x}, \mathbf{W}) = |\hat y - y_c|^{-1}$ to ensure a close distance is proportional to high confidence, and normalized over $\mathcal{C}$.
The deep model depicted in Fig.~\ref{fig:architecture} was trimmed to preserve the $f$ model in order to generate only $\hat y$.

\subsubsection{MC-Dropout}
The MC-Dropout method is an approximate Bayesian inference method that can capture the epistemic uncertainty.
Similarly to the regression baseline method, the model depicted in Fig.~\ref{fig:architecture} was trimmed down to only the $f$ model.
Then, we add a \emph{permanent} Dropout layer before every layer with trainable parameters to make sure the Dropout is also functional during the inference stage.
The drop rate is set to 0.1 uniformly for all Dropout layers.

Note the off-the-shelf DenseNet and many other state-of-the-art deep nets integrate Dropout to prevent over-fitting, which to an extent reduces the epistemic uncertainty. 
Our intention of including this method is to verify how effective it can be by having the extra Dropouts at a cost of a slower convergence and a higher test time complexity.
The confidence is also approximated by the aforementioned inverse distance method.

\subsubsection{PDF-Prob} We denote the third method PDF-Prob as it uses Gaussian PDF approximated density (see Eq.~(\ref{eq:pdf})), where the training relies on the regression label $y$~\cite{nix1994estimating}.
The density is normalized to relative likelihood probability for computing the confidence.
For the PDF-Prob method, the $f$ and $g$ models in Fig.~\ref{fig:architecture} are all preserved; however, the CDF-Prob layer was removed and the estimated $\hat y$ and $\hat \sigma$ were supplied directly to the loss function shown in Eq.~(\ref{eq:aleatoric_regression}) to compute the gradient. 

\subsubsection{CDF-Prob} CDF-Prob denotes our proposed enhancement over the PDF-Prob method to appropriately handle the categorical label.
The CDF-Prob method is also combined with Laplace density for comparison.

\subsection{Comparison and Discussion}

\begin{figure*}
    \centering
    \resizebox{\textwidth}{!}{%
    \begin{tabular}{c}
        \includegraphics[width=1.0\textwidth]{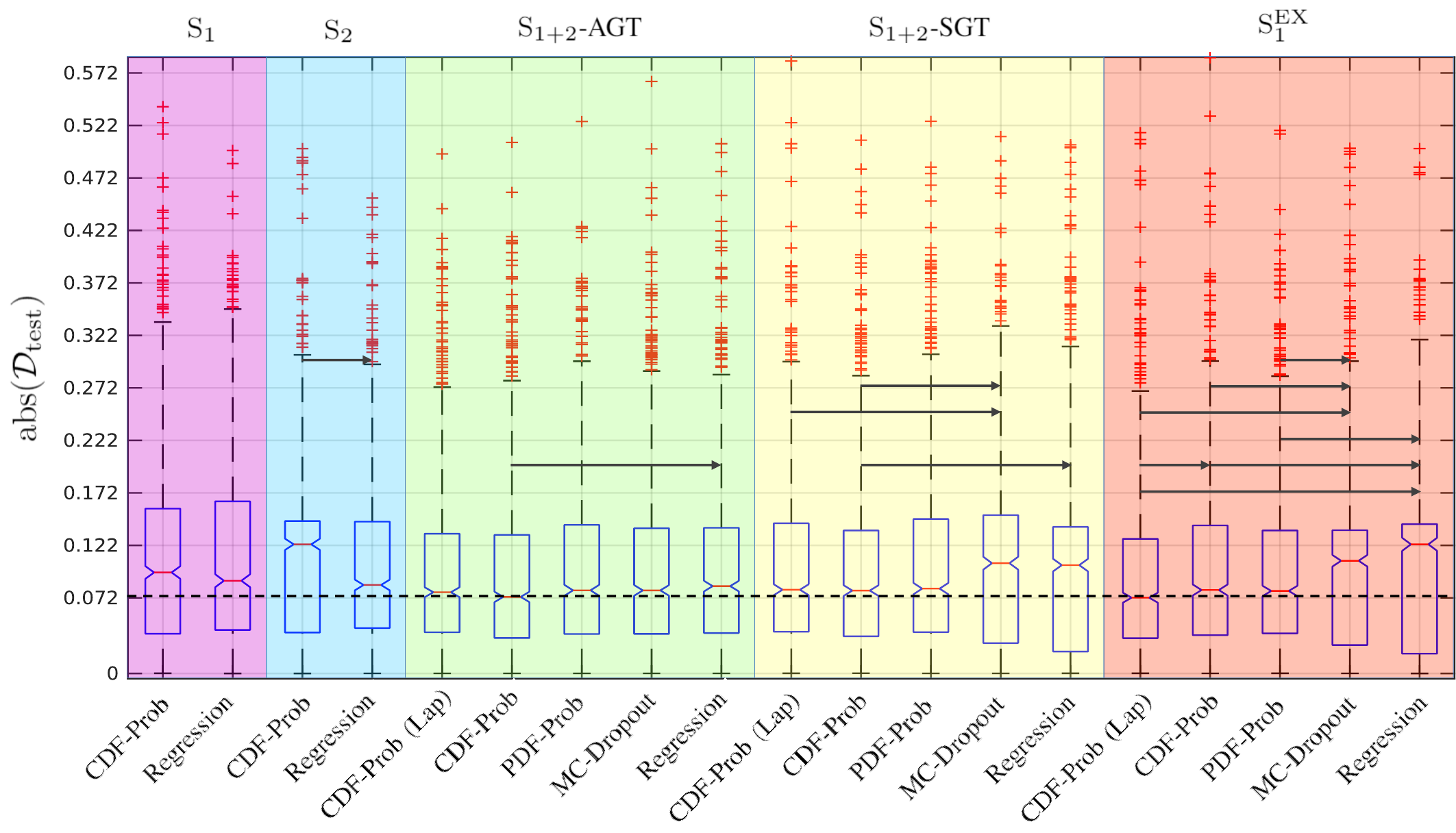} \\
    \end{tabular}%
    }
    \caption{The absolute error distribution boxplot comparison of the results listed in Table~\ref{tb:compare_methods_ensemble_performance} in App.~\ref{app:tables}.
    The globally observed minimal median absolute error (0.072) is shown as a dotted line.
    Within each group, the arrows indicate the mean absolute error of the dull side method is statistically significant smaller than that of the pointy side, computed by one tail hypothesis using  two-sample $t$-test.}
    \label{fig:boxplot}
\end{figure*}

We evaluate the performance of the aforementioned methods by an ensemble of five repetitively trained model instances, where each training instance was initialized with a different random seed.
The numerical results can be found in Table~\ref{tb:compare_methods_ensemble_performance} in App.~\ref{app:tables} and the comparison of $\mathrm{abs}(\mathcal{D}_\mathrm{test})$ distributions are graphically presented in Fig.~\ref{fig:boxplot}.
A low absolute error and a high classification accuracy are considered as \emph{surrogate indicators} of successful modelling of intra-observer variability information. 
In Table~\ref{tb:significance_table} in the App.~\ref{app:tables}, the listed models have been tested by one tail hypothesis with the use of two-sample $t$-test in order to justify the improvements discussed in the following sections.

\subsubsection{General observations}
it can be observed that the classification accuracy of the uncertain studies are saturated over 90\% due to the relaxed accuracy formula (see Eq.~(\ref{eq:accu})), while the accuracy of the certain studies are between $61\%\sim73\%$. 
Another observation is that the certain studies have 22\% more population in $\mathcal{D}_\mathrm{test}$ compared to the uncertain studies, suggesting the obtained results on the entire test set is slightly more biased to the performance of the certain studies.
Furthermore, an approximately 0.02 gap between each pair of the mean and median absolute errors indicate the presence of outliers with large absolute errors.
In addition, we observe a disconnection between the accuracy measure and the calibration metrics, \ie, a higher accuracy does not necessarily bind a lower ECE or MCE value.
This phenomenon can be explained as a disconnection between negative log-likelihood (NLL) training and accuracy metric as neural nets can overfit NLL without overfitting to the 0/1 loss~\cite{guo2017calibration}.
In our opinion, the confidence is directly affected by the underlying density distribution, which also explains the ECE drop-off from CDF-Prob models to CDF-Prob (Lap) models.

\subsubsection{Soft targets and LDL methods pre-code observer uncertainty}
\label{sec:soft_targets}

it is noticeable that the $\mathrm{S_{1+2}}$-AGT trained models have most competitive performance based on the following observations:
\begin{itemize}
    \item only in $\mathrm{S_{1+2}}$-AGT, every model significantly improves over the baseline $\mathrm{S_\#}$ trained regressions with 95\% confidence interval (see
the 2$^\text{nd}$ \& 4$^\text{th}$ columns of Table~\ref{tb:significance_table} in App.~\ref{app:tables});
    \item in particular, the $\mathrm{S_{1+2}}$-AGT trained regression model can be translated to 81.6\% in accuracy, improves over $\mathrm{S_{\#}}$ trained regressions by 4.2\% and 5\%, respectively.
\end{itemize} 
It is worth mentioning the second point as the improvement is achieved without the help of  uncertainty modelling techniques.

\subsubsection{Stability is much more helpful than exploratory ability}
the evidences come from the AGT and SGT model comparisons: 
\begin{itemize}
    \item for the MC-Dropout model, the improvement from SGT to AGT is significant with 95\% confidence interval, translating to 4.5\% accuracy increase;
    \item for the regression model, the improvement is significant with 90\% confidence interval, translating to 3.2\% accuracy increase;
    \item for the PDF-Prob model, the improvement is significant with 90\% confidence interval, translating to 1.5\% accuracy increase;
    \item finally, for CDF-Prob and CDF-Prob (Lap), the accuracy increases are 1.2\% and 0.7\%, respectively, but the mean absolute errors fail to show statistically significantly improvement.
\end{itemize}
The conclusion that the stability in AGT is more helpful to reduce observer uncertainty compared to the exploratory ability of SGT is understandable, as the exploratory ability introduces extra uncertainty (in the form of the random label flipping noise).
The gradually closing performance gap from the first to last point indicates that the proposed uncertainty modelling methods can handle the random uncertainty introduced by SGT and mitigate the performance gap.

\subsubsection{``More labels? More data?''}
\label{sec:more_what}
data acquisition usually imposes a huge overhead to the medical imaging research.
Compared to using archived data, the prospective acquisition is even more time consuming.
Hence, acquiring multiple rounds of expert labels may render a much more economic solution.
It is worth to investigate whether acquiring more labels is as beneficial as acquiring more data.

\begin{figure*}[!htbp]
    \centering
    \resizebox{0.8\textwidth}{!}{%
    \begin{tabular}{cc}
        \includegraphics[width=0.5\textwidth]{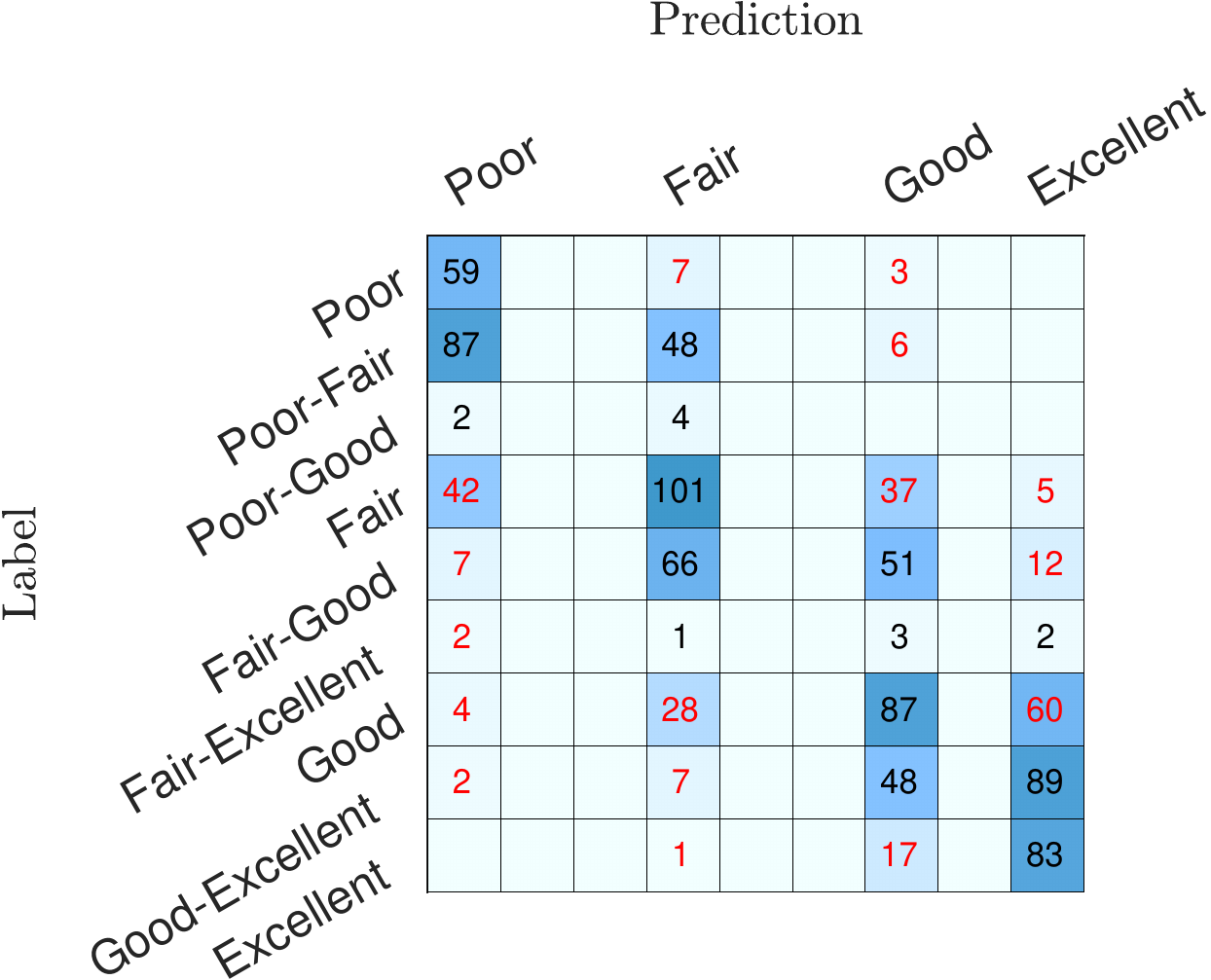} & 
        \includegraphics[width=0.5\textwidth]{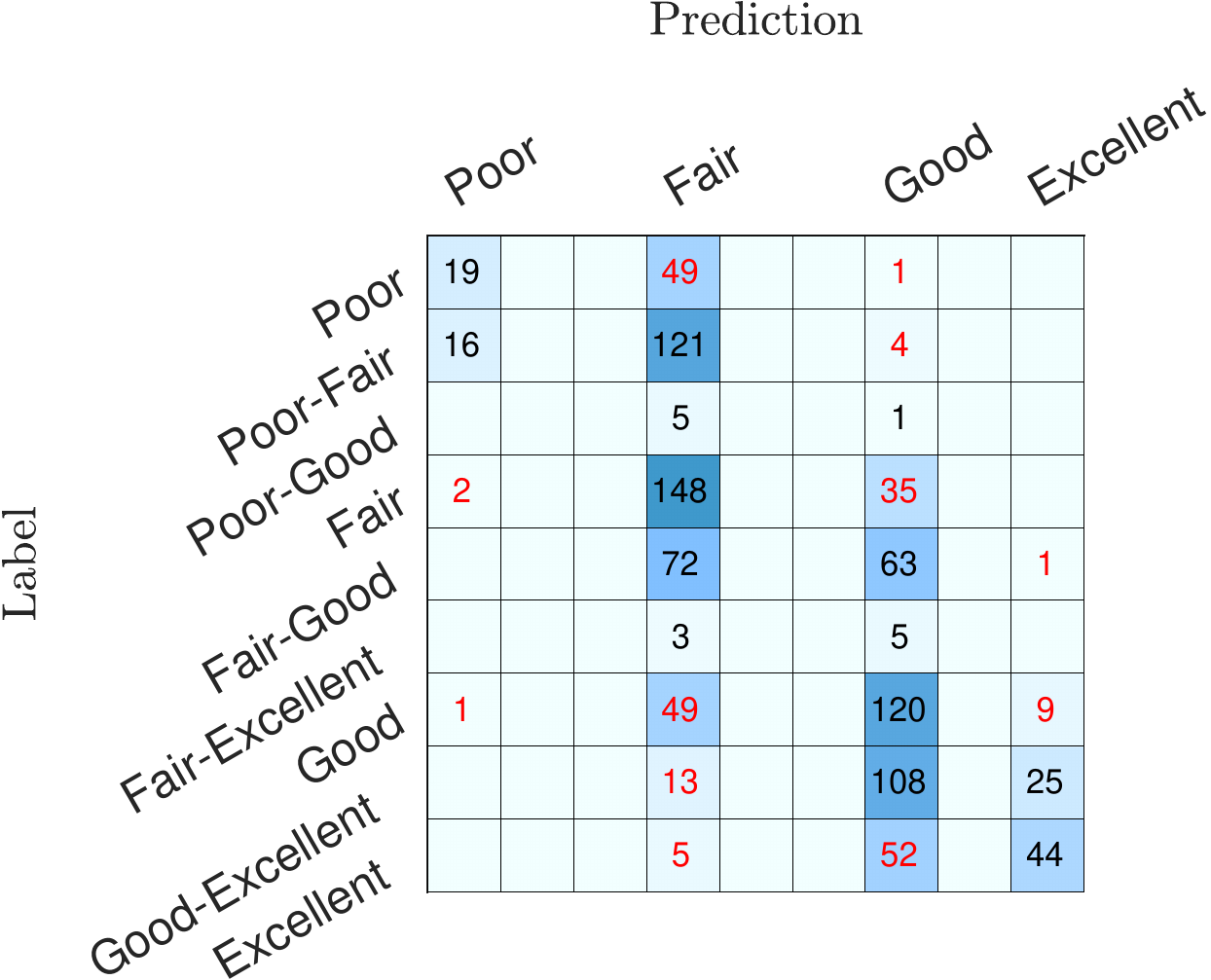} \\ 
        (a) $\mathrm{S_1}$  & (b) $\mathrm{S_2}$ \\
        \\
        \includegraphics[width=0.5\textwidth]{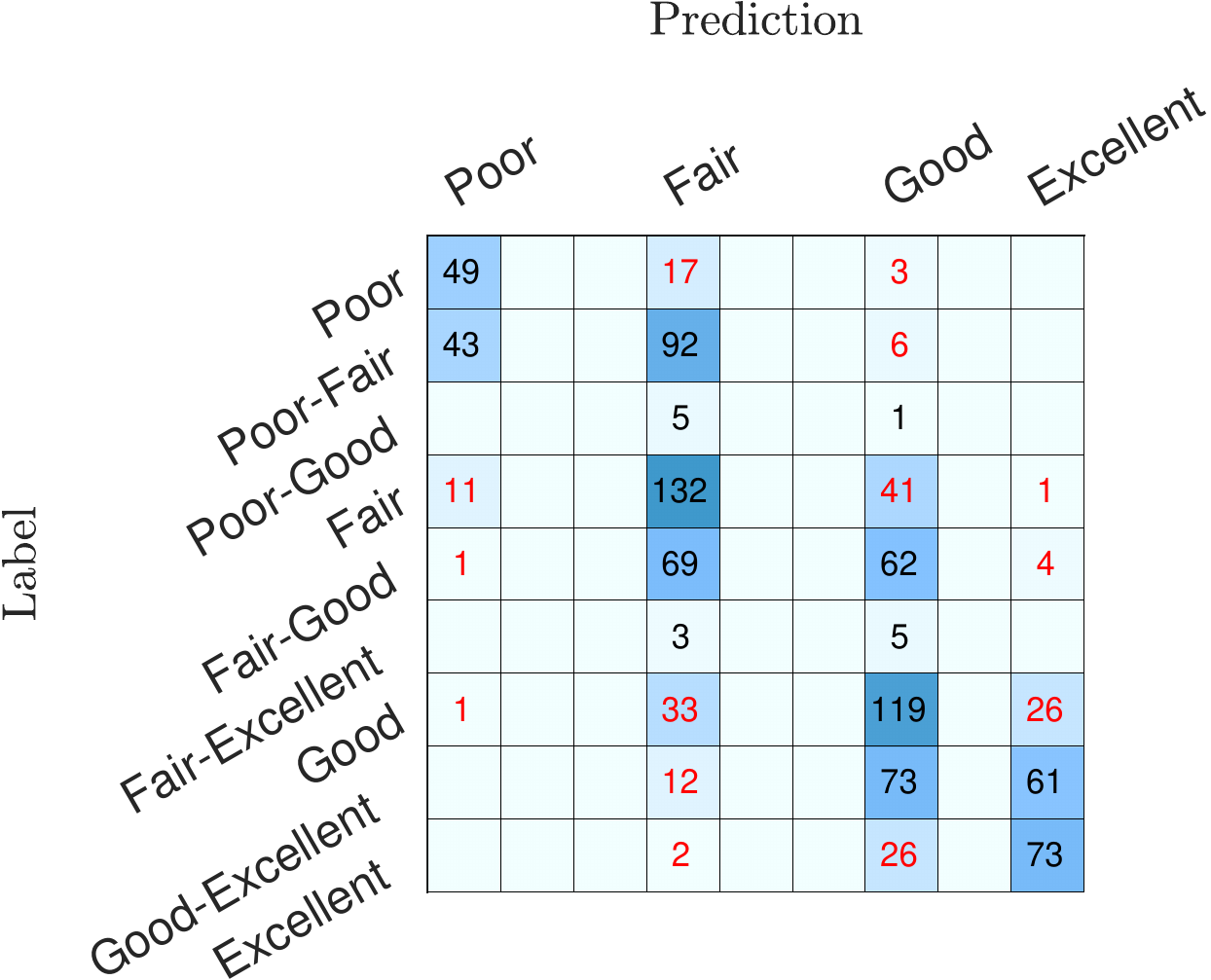} & 
        \includegraphics[width=0.5\textwidth]{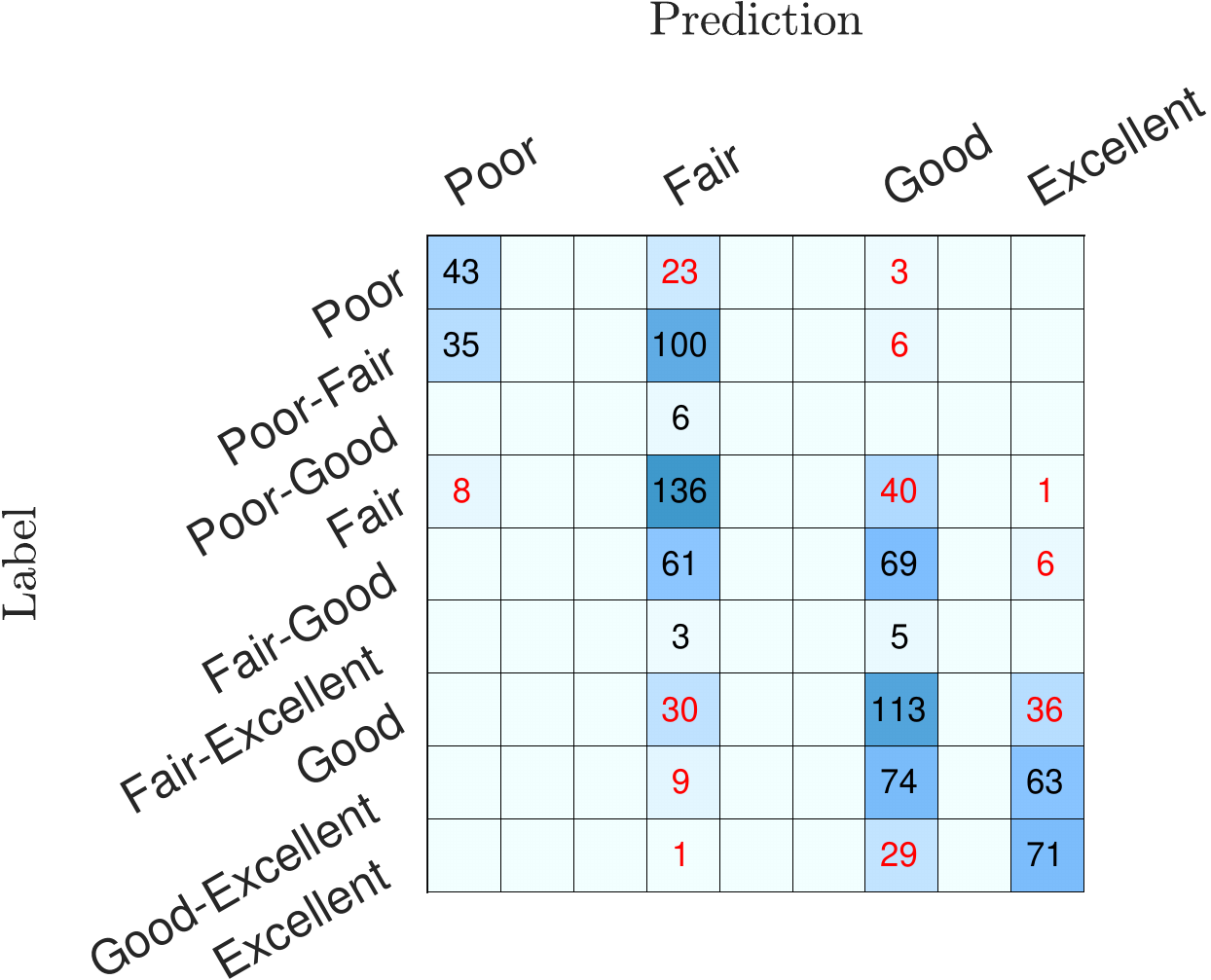} \\
        (c) $\mathrm{S_{1+2}}$-AGT & (d) $\mathrm{S_{1+2}}$-SGT\\
    \end{tabular}%
    }
    \caption{The confusion matrices for the classification performance of the CDF-Prob method trained under four different scenarios. 
    In addition to the four confident classes, the Label axis also shows the five unconfident classes. 
    However, the Prediction axis only shows the four confident classes as we only show the categorization of the expectation $\hat \mu$ being one of the four pre-defined classes.
    The red colored numbers indicate the incorrect classification.}
    \label{fig:four_confusion_matrices}
\end{figure*}

First of all, acquiring more labels is beneficial.
The evidences are:
\begin{itemize}
    \item the aforementioned improvement from $\mathrm{S_{1+2}}$-AGT trained regression model to $\mathrm{S_\#}$ trained regressions (see the 2$^\text{nd}$ point in Sec.~\ref{sec:soft_targets});
    \item $\mathrm{S_{1+2}}$-AGT trained CDF-Prob model is also statistically significantly better than the respective $\mathrm{S_\#}$ trained CDF-Prob models; suggesting that a standalone $\mathbf{A_\#}$ set does not contain the same level of uncertainty information as contained by the combination of both sets.
\end{itemize}
We show the confusion matrices of the CDF-Prob method under the two $\mathrm{S_\#}$ and the two $\mathrm{S_{1+2}}$ scenarios in Fig.~\ref{fig:four_confusion_matrices}.
It can be observed that the CDF-Prob model trained with $\mathrm{S_1}$ is likely to confuse the two middle classes with the two boundary classes while the CDF-Prob model trained with $\mathrm{S_2}$ shows the opposite; on the other hand, the CDF-Prob models trained under the two $\mathrm{S_{1+2}}$ scenarios alleviate such confusion.

Next, acquiring more data is also beneficial but a proper uncertainty modelling is necessary to realize the improvement.
The evidences are:
\begin{itemize}
    \item the performance of $\mathrm{S_{1}^{EX}}$ trained regression model is at the same level of the $\mathrm{S_{\#}}$ trained regressions, which does not justify the collection cost of the additional data volume and the prolonged training time to iterate through the dataset;
    \item $\mathrm{S_{1}^{EX}}$ trained uncertainty modelling models CDF-Prob, CDF-Prob (Lap), and PDF-Prob, which all obtained statistically significant improvements over $\mathrm{S_{\#}}$ trained regression models.
\end{itemize}

In short, the decision between more labels or more data is in favour towards more labels, as $\mathrm{S_{1+2}}$-AGT trained CDF-Prob model is significantly better than the $\mathrm{S_{1}^{EX}}$ trained counterpart with 90\% confidence interval.
In conclusion, given limited resources to acquire new data, acquiring more labels can be beneficial and much more economical than acquiring more image data (which would also require labelling the acquired data). Note that active learning methods can be applied to further reduce the acquisition and labelling cost by selecting informative samples.

\begin{figure*}
    \centering
    \resizebox{0.85\textwidth}{!}{%
    \begin{tabular}{cc}
    \includegraphics[width=0.425\textwidth]{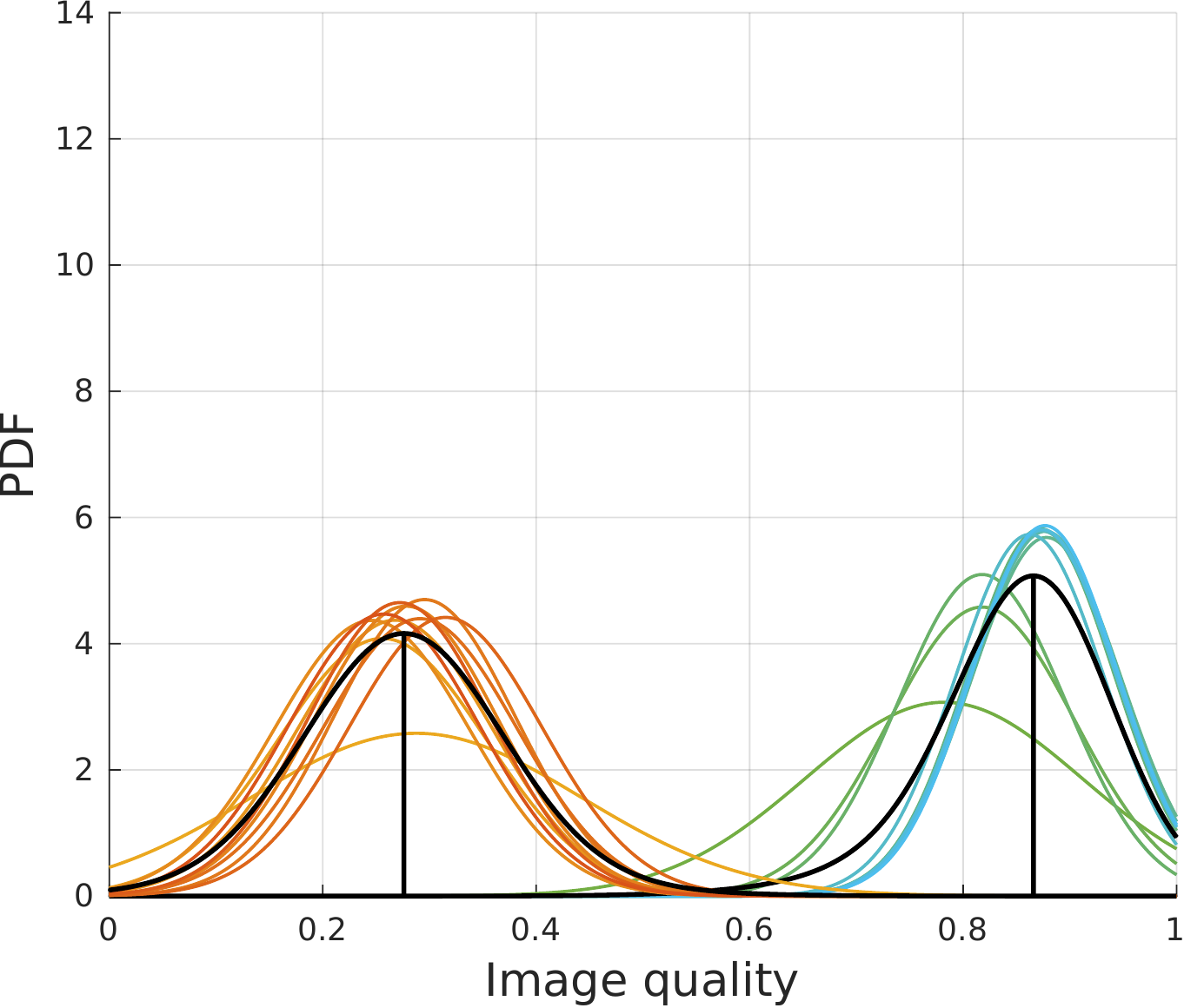} &
    \includegraphics[width=0.425\textwidth]{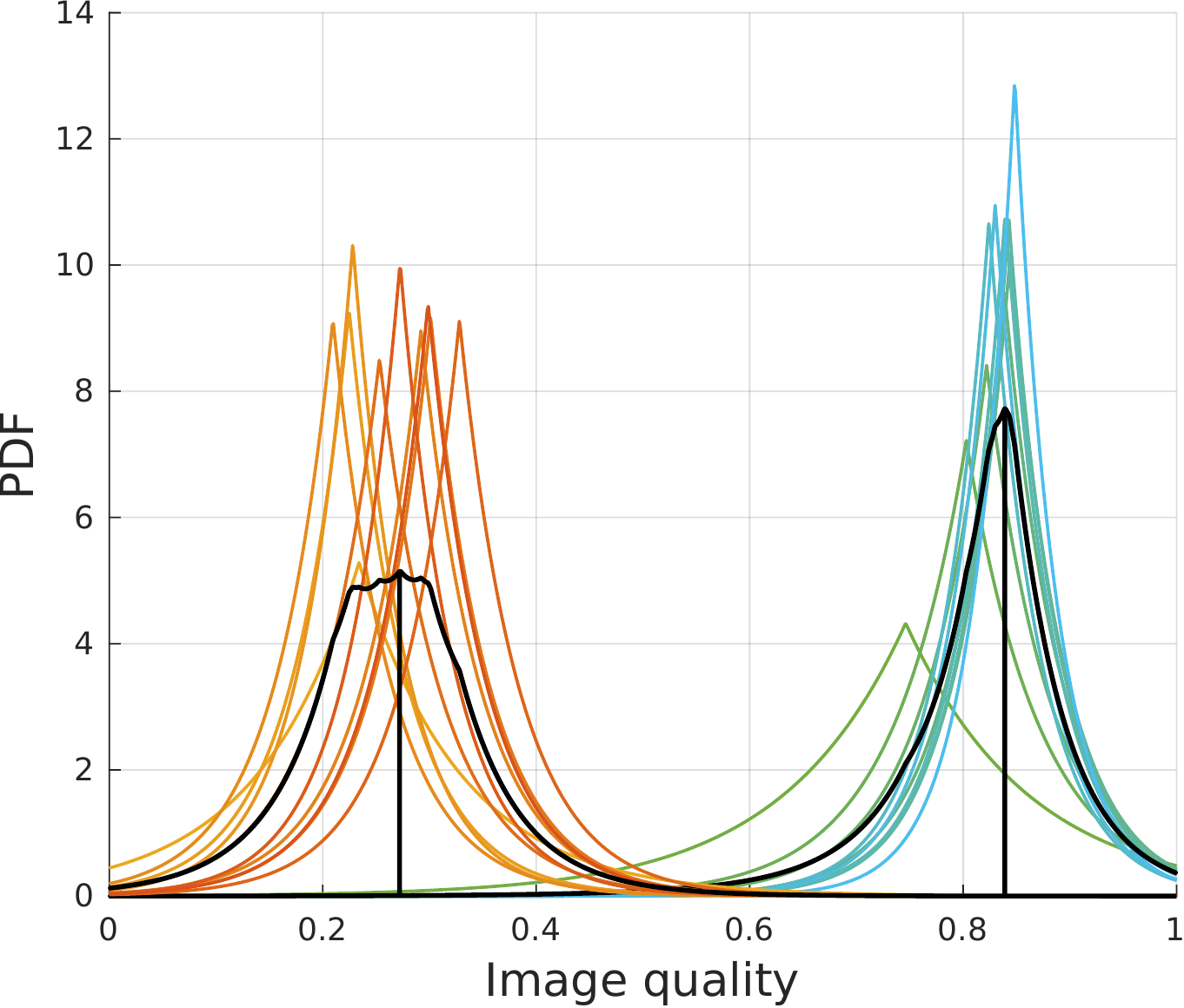} \\ 
    (a) CDF-Prob Mixture & (b) CDF-Prob Mixture (Lap)\\
    \end{tabular}%
    }
    \caption{The PDF plots of the estimated mixture models (in black color, the vertical line marks the expected image quality) and their sub 10 mixture components for a low quality sample (in yellow to orange colors, marking the first step to the last step) and a high quality sample (green to blue color, also marking the first step to the last step). }
    \label{fig:pdfs}
\end{figure*}

\subsubsection{Modelling intra-observer variability as aleatoric uncertainty}
the DenseNet architecture itself has integrated Dropout layers, which were not removed in our experiments in order to determine the baseline performance of the original DenseNet proposal, meaning the epistemic uncertainty is modelled to some degree in the respective regression models.
To properly handle the epistemic uncertainty, we tested full MC-Dropout in the respective scenarios.
Since both AGT and SGT have the ability to reduce intra-observer variability which will be reflected on the mean absolute error and accuracy, the appropriate comparison between MC-Dropout and regression is on $\mathbf{S_{1}^{EX}}$, where we observe the ``full dropout'' does not necessarily provide improvement over the regression model, yet it introduces significant training complexity (in our experiment setting, this incurs 33\% more training time on average).

Hence, the CDF-Prob (Lap), CDF-Prob, and PDF-Prob methods were built on the original DenseNet for maintaining a lower training and test complexity, where we assume the default Dropout layers are sufficient to handle the epistemic uncertainty of the tested model.
Without the help from the AGT or SGT, CDF-Prob (Lap), CDF-Prob, and PDF-Prob methods are able to obtain statistically significant improvements over regression under the $\mathbf{S_{1}^{EX}}$ scenario, translating to 5.7\%, 4\%, and 3.2\% improvements in accuracy.
In association with the marginal improvement of the MC-Dropout regression, we advocate that the intra-observer variability (at least the proportion that has been captured) were modelled as aleatoric uncertainty.
In fact, the experiments in $\mathbf{S_{1}^{EX}}$ scenario also indicate that the proposed CDF-Prob method does not necessarily need extra labeling to model the aleatoric uncertainty.

In our opinion, the advantage of AGT (\ie, soft targets and LDL) is to pre-code the uncertainty in the labels so that a generic learning model can be applied without concerning about modelling the uncertainty, while the advantage of the proposed PDF-Prob and CDF-Prob is that when we do not have the exact many-to-one data-to-label correspondence, the uncertainty can be appropriately modelled.

\subsubsection{CDF-Prob Laplace variation}

the modification to replace the Gaussian density with the Laplace distribution happens inside the CDF-Prob layer by editing the $F(.)$ definition.
Comparing only on the numerical values, we observe the optimal performance is with Gaussian distribution in the $\mathrm{S_{1+2}}$ scenarios but with Laplace distribution in $\mathrm{S_1^{EX}}$ scenario.
It is intuitive for the boundary classes \emph{Excellent} and \emph{Poor} that Gamma distribution can be a better approximation.

\begin{figure}
    \centering
    \includegraphics[width=1\columnwidth]{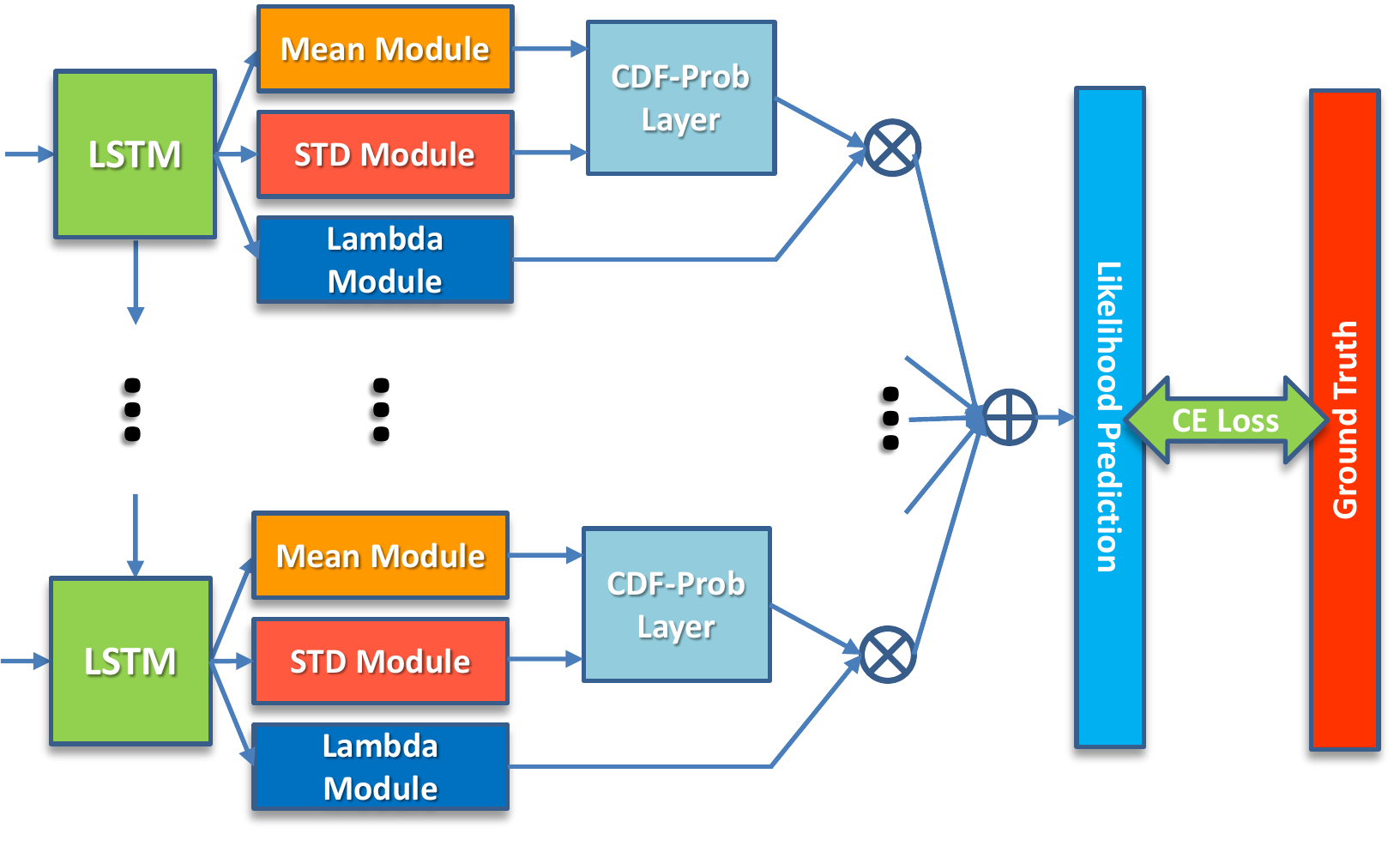}
    \caption{Schematic of the mixture model architecture, only the components that differ from Fig.~\ref{fig:architecture} are shown.}
    \label{fig:mixture_architecture}
\end{figure}

\subsubsection{Mixture model variation}

for demonstrating the compatibility of the proposed CDF-Prob in a mixture model setting, we introduce a 10-component mixture which can be found in Fig.~\ref{fig:mixture_architecture}.
This model shares the same DenseNet + LSTM base model to the model defined in Fig.~\ref{fig:architecture}.
The reason of having 10 components is because of the 10 time steps, where each step does have $\hat \mu$ and $\hat \sigma$ estimations before the averaging.
Hence, the mixture model architecture has the ``Averaging Mean'' and ``Averaging STD'' modules removed, and the per step $\hat \mu$ and $\hat \sigma$ estimations are directly plugged to one CDF-Prob module to estimate $\hat e$.
In parallel to the Mean and STD modules, the LSTM features are also fed to compute an additional value lambda (softmaxed over the 10 steps) to be the weighting parameter for each component.
The weighted CDFs are then summed to compute the final likelihood distribution, which replaces the usage of Eq.~(\ref{eq:true_distribution}) for training the model.
Such learned softmax weighting parameters share the concept of attention mechanism.

We tested the mixture model with both Gaussian and Laplace distributions, namely the CDF-Prob Mixture and CDF-Prob Mixture (Lap) models, under the $\mathrm{S_1^{\mathrm{EX}}}$ scenario, resulting to $83.3\%$ and $82.8\%$ in accuracy and $0.094 \pm 0.081\;(0.074)$ and $0.095 \pm 0.081\;(0.075)$ in mean$\pm$std (median) absolute errors, respectively.
Comparing the respective non-mixture model counterparts, the Gaussian mixture model shows significant improvement but the Laplace models perform similarly.
In Fig.~\ref{fig:pdfs}, we show the mixture model PDF and the 10 individual components' PDFs for two samples.
It can be noticed that the first few steps intend to have lower image quality expectations and larger STDs while the latter steps have tighter distributions with higher quality expectations.

\begin{figure*}[!htbp]
    \centering
    \resizebox{0.9\textwidth}{!}{%
    \begin{tabular}{ccc}
    \includegraphics[width=0.448\textwidth]{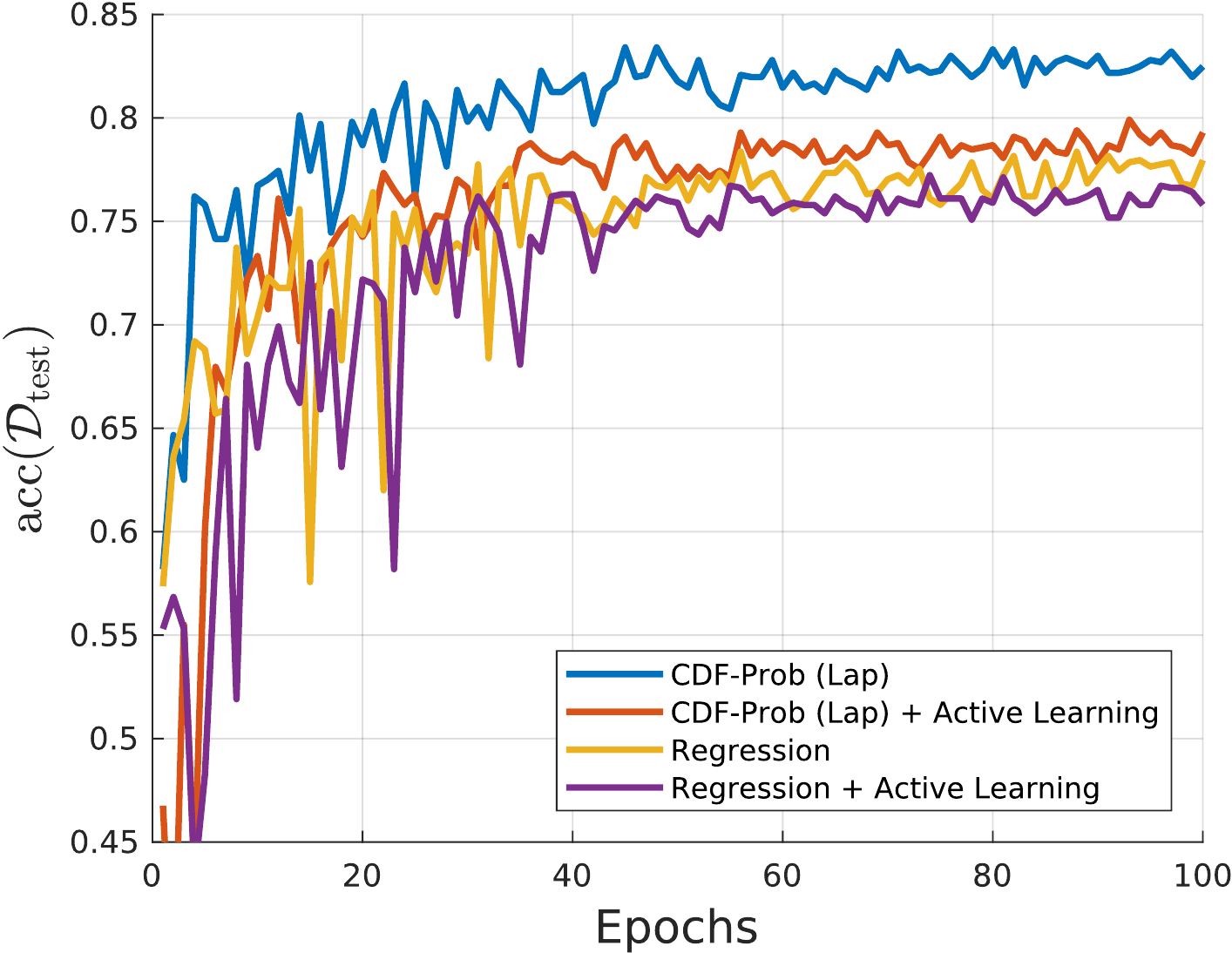} & 
    \includegraphics[width=0.44\textwidth]{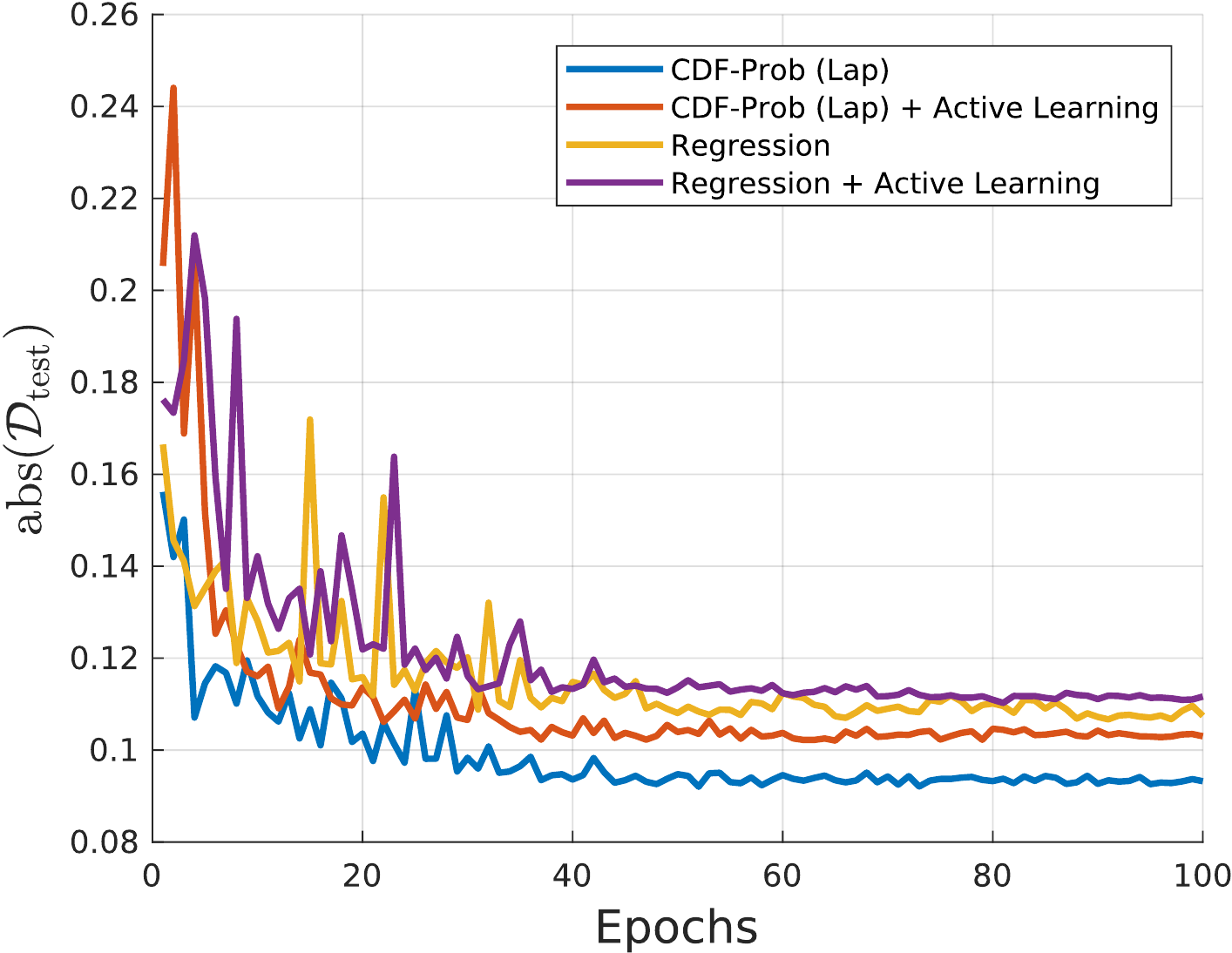} \\
    (a) accuracy curves &
    (b) mean absolute error curves & 
    \end{tabular}%
    }
    \caption{The test accuracy (a) and the test mean absolute error (b) curves recorded during the training of the respective shown models.}
    \label{fig:active_learning}
\end{figure*}

\begin{figure}
    \centering
    \includegraphics[width=0.9\columnwidth]{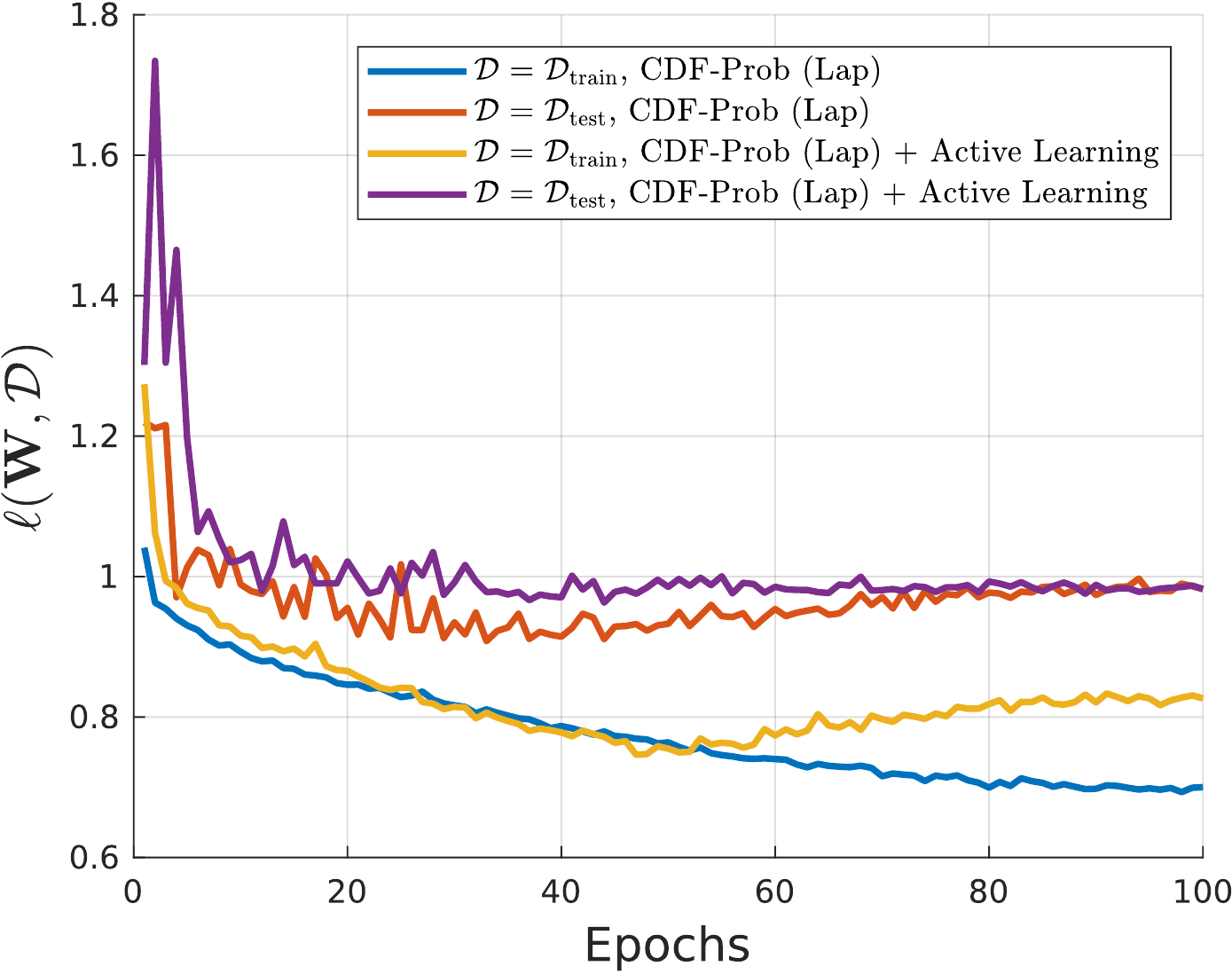}
    \caption{The loss curves of CDF-Prob (Lap) with and without active learning. }
    \label{fig:active_learning_loss}
\end{figure}

\subsubsection{Combining with Active Learning} active learning is aimed at reducing the labelling cost.
We show that our proposed model also functions well in an active learning setting.
The details about the setup can be found in App.~\ref{app:active_learning}.
An uncertainty sampling learner model~\cite{lewis1994heterogeneous} treating low confidence of the prediction, \ie, $\mathrm{conf}$, as the measure for hard samples is used in this experiment.

The main result is illustrated in Fig.~\ref{fig:active_learning}, where we found the tested active learning trained models, namely the Regression + Active Learning and the CDF-Prob (Lap) + Active Learning models, do not perform better than their respective full set trained counterparts.
This may suggest that the left out ``more certain'' samples may also be considered informative to model the underlying expert distribution $e$.
Nevertheless, the proposed CDF-Prob (Lap) method does show a consistent improvement over regression.

In Fig.~\ref{fig:active_learning_loss}, we include the loss convergence graphs of the CDF-Prob (Lap) method with and without active learning, where we can observe after the 50$^\text{th}$ epoch (where the active learning algorithm actually starts), learning with the most uncertain samples incurs convergence difficulties.

\subsubsection{Classification} finally, we test the classification approach.
Our classification model uses the same DenseNet + LSTM model but with a fully-connected output layer, activated by softmax to compute the categorical quality predictions at each time step.
The individual predictions are averaged for direct estimation of $\hat p_c$. 
This model is shown as ``Classification'' in Table~\ref{tb:kld_cls}.
It is noticeable that the classification accuracy is 8.5\% lower than the proposed CDF-Prob method.
On $\mathrm{S_{1+2}}$-AGT, the classification method scored 77.2\%, where all regression based methods were able to achieve accuracy above 81\%.
The main reason we choose regression instead of classification in the main experiment is to explicitly project the quality on a 1-D output space.
This imposes a class-wise geometrical order in the output space, \ie, the \textit{Poor} and \textit{Good} classes should be on opposite sides of the the \textit{Fair} class.
On the other hand, classification has a 4-D output space corresponding to the number of the classes; hence it is possible that the classes reside on a curved manifold embedding so that the distance between the \textit{Poor} and \textit{Good} classes may be shorter than the distance measured from the \textit{Poor} class to the \textit{Fair} class then to the \textit{Good} class.
In our opinion, the lower accuracy of the classification on $\mathrm{S_{1+2}}$-AGT could be a result of such freedom in the output space as the soft targets of a confused study with \textit{Poor} and \textit{Good} labels is an indicator that these two classes may be closer to each other than each to the rest of classes.

\begin{table}[!htbp]
    \centering
     \caption{The accuracy comparison between the CDF-Prob, PDF-Prob and classification methods.}
    \resizebox{\columnwidth}{!}{%
    \begin{tabular}{|l|c|c|c|c|c|}\hline
													
\multicolumn{6}{|c|}{acc (\%)}											\\
\hline	&	$\mathrm{S_1}$	&	$\mathrm{S_2}$	&	$\mathrm{S_{1+2}}$-AGT	&	$\mathrm{S_{1+2}}$-SGT	&	$\mathrm{S_1^{\mathrm{EX}}}$	\\
\hline											
CDF-Prob (Lap)	&	-	&	-	&	$81.9$ 	&	$80.7$	&	$\bf 82.8$	\\
\hline											
CDF-Prob	&	$77.0$	&	$79.7$	&	$\bf 82.9$	&	$\bf 82.2$	&	$81.1$	\\
\hline											
PDF-Prob	&	-	&	-	&	$81.4$	&	$79.9$	&	$80.3$	\\
\hline											
Classification	&	$68.5$	&	$77.5$	&	$77.2$ 	&	$79.3$	&	$80.6$	\\
\hline																																																						
    \end{tabular}%
    }
   
    \label{tb:kld_cls}
\end{table}

\section{Conclusion}

In this paper, we propose a method for modelling label uncertainty in deep neural networks. We demonstrate the efficacy of the approach in the context of echocardiography quality assessment as an important step towards accurate diagnosis of cardiovascular disease. Two specific challenges associated with data labels have been addressed: 1) there is large intra-observer variability in data labels; and 2) there is a need to solve a regression problem in the presence of categorical labels.
Under the tested scenarios, we found the CDP-Prob method marginally improves over PDF-Prob by the absolute error measurement, which equivalents to 0.8\%-2.5\% improvement in accuracy.
Compared to the plain regression method, CDF-Prob shows statistically significant improvement in respective scenarios, equivalent to 1.3\% to 5.7\% improvement in accuracy, but PDF-Prob model does not.

In a hypothetical scenario where more data are needed, relabelling the same dataset and leveraging soft labels can be the most beneficial, economical and simplest solution to reduce the effect of observer variability;
in another hypothetical scenario when there is sufficient data but without multiple labels, an aleatoric uncertainty modelling method, such as the PDF-Prob or CDF-Prob, alone could be a helpful solution.

Compared to the stochastic methods for uncertainty estimation, which require stochastically sampling the network model multiple times for estimating the uncertainty (\eg, MC-Dropout), both PDF-Prob and CDF-Prob parameterize the uncertainty distribution; hence a single forward pass is sufficient to estimate the level of uncertainty as the STD value. 

The main limitation of the study is with the limited availability of expert labelling, which allows a coarse performance analysis for the compared methods. 
Furthermore, it also limits the comparison against the widely used \emph{majority voting} composition.
Essentially, the CDF-Prob trained model is a classification approach with a slightly
altered
\emph{Softmax} layer, incurring negligible cost to training complexity,
hence it works similar to a plain classification model. 

Finally, we want to reiterate that the advantage of using the proposed CDF-Prob method is the explicitly computed numeral assessment values even trained with the categorical labels. 
Our formulation is generic and the approach can be applied in scenarios where there is variability in expert labels or where there is a high discrepancy in clinical measurements. 

\bibliographystyle{IEEEtran}
\bibliography{mybib}

\begin{appendices} 

\section{Gaudet chart} 
\label{app:gaudet}
The Gaudet chart~\cite{gaudet2016focused} is a point-based scoring system assessing the echo image quality based on several sub-criteria: 
\begin{itemize}
\item presence and contrast of target heart anatomies (\ie, valve and chamber wall, and these are view dependent);
\item proper centering;
\item imaging depth setting;
\item imaging gain setting.
\end{itemize}
As a result, the Gaudet score system is a time consuming marking process.
As an example, the AP4 and AP2 view charts are shown in Table~\ref{tb:gaudet}.

\begin{table}[!htbp]
    \caption{The Gaudet marking charts for AP4 and AP2 quality assessment.
    The increment for each sub critira is 0.5.}
    \centering
    \begin{tabular}{|l|c|}
    \hline
    \bf{AP4} & \bf{Score} \\
    \hline
    Left and right ventricles & /2 \\
    \hline
    Left and right atriums & /2 \\
    \hline
    Mitral and tricuspid valves visible & /2 \\
    \hline
    Mitral and tricuspid values optimized & /2 \\
    \hline
    Structures centered in view & /1 \\
    \hline
    Appropriate depth & /0.5 \\
    \hline
    Appropriate gain & /0.5 \\
    \hline
    {\bf TOTAL SCORE} & {\bf /10} \\
    \hline \hline
    \bf{AP2} & \bf{Score} \\
    \hline
    Left ventricle & /2 \\
    \hline
    Left atrium & /2 \\
    \hline
    Mitral valve visible & /2 \\
    \hline
    Structures centered in view & /1 \\
    \hline
    Appropriate depth & /0.5 \\
    \hline
    Appropriate gain & /0.5 \\
    \hline
    {\bf TOTAL SCORE} & {\bf /8} \\
    \hline
    \end{tabular}
    
    \label{tb:gaudet}
\end{table}

\section{Determination of the Four-category subjective labelling scheme and Labelling Guideline} 
\label{app:4class}
The four-category labelling scheme was derived from the Gaudet point system.
Given that different views have different maximum Gaudet scores, the subjective labelling score range is normalized between 0 and 1 for each view, and quantified into the equally ranged four categories from \emph{Poor} to \emph{Excellent}.
Note that the study cine series are real hospital data acquired by qualified sonographers. 
Hence, after further discussions with clinicians, we did not consider the class \emph{Unacceptable} (0\%), \eg, non-cardiac cine series, as described by Abdi \etal~\cite{abdi2017automatic}.
As for the reason of having low quality data in the collected dataset, it may be related to the possible atypical probe position during acquisition, which is due to some clinical conditions, especially post cardiac and thoracic surgeries and tapping of pericardial  effusion, where big area of the chest can be covered with adhesive plaster. 
In some other cases, patients have poor parasternal view acquisition window, especially patient with emphysema, chronic obstructive pulmonary disease (COPD) and chest deforestation, where sonographers have to use different probe positions. 
Patients on mechanical ventilation are usually scanned on supine position which might have lower image quality than that of left lateral position.
In addition, some patients with severe dyspnoea or orthopnoea are usually scanned while sitting, and this might have lower image quality.

During the labelling process, the senior cardiologist only has access to the B-mode image data extracted from the DICOM file on a PC monitor, where the meta information (\eg, Ultrasound machine information and patient history) were not displayed.
Example images from each category were provided before the labelling trials to help him comprehend the concept of each quality category.
Although the cardiologist was well aware of the Gaudet scoring system, the chart was provided to him during the labelling process for reference.
For cine series with variable quality, the score was dependent on the frames with the best quality since these frames will be used to analyze the heart function.

\begin{table*}[!htbp]
    \centering
     \caption{The accuracy (see Eq.~(\ref{eq:accu})) and mean absolute error (see Eq.~(\ref{eq:abs}), shown also with the STD and median) comparison of methods based on an ensemble of five repetitively trained instances for each method.
    The measurements are also computed for the certain and uncertain study groups (see the respective definitions at the end of Sec.~\ref{sec:dataset}).}
    \resizebox{\textwidth}{!}{%
    \begin{tabular}{|l|c|c|c|c|c|c|c|c|c|c|}
    \hline
    \multirow{2}{*}{$\mathrm{S_1}$}	&	\multicolumn{4}{c|}{Entire Test Set}							&	\multicolumn{2}{c|}{Certain Test Studies (55\%)}							&	\multicolumn{2}{c|}{Uncertain Test Studies  (45\%)}							\\
\cline{2-9}																									
	&	acc (\%)	&	abs: mean $\pm$ std (median)	&	ece (\%)	&	mce (\%)	&	acc (\%)	&	abs: mean $\pm$ std (median)					&	acc (\%)	&	abs: mean $\pm$ std (median)					\\
\hline																									
CDF-Prob	&	$77.0$ & $0.114 \pm 0.094 \; (0.088)$ & $21.7$ & $65.8$ &$65.2$ & $0.112 \pm 0.093 \; (0.083)$ & $91.5$  & $0.094 \pm 0.094 \; (0.116)$ \\																							
\hline																									
Regression	&	$77.4$ & $0.112 \pm 0.096 \; (0.096)$ & $15.8$ & $40.9$ &$65.5$ & $0.107 \pm 0.100 \; (0.085)$ & $92.0$  & $0.108 \pm 0.091 \; (0.119)$ \\																							
\hline \hline

\multirow{2}{*}{$\mathrm{S_2}$}	&	\multicolumn{4}{c|}{Entire Test Set}							&	\multicolumn{2}{c|}{Certain Test Studies (55\%)}							&	\multicolumn{2}{c|}{Uncertain Test Studies  (45\%)}							\\
\cline{2-9}																									
	&	acc (\%)	&	abs: mean $\pm$ std (median)	&	ece (\%)	&	mce (\%)	&	acc (\%)	&	abs: mean $\pm$ std (median)					&	acc (\%)	&	abs: mean $\pm$ std (median)					\\
\hline																									
CDF-Prob	&	$79.7$ & $0.102 \pm 0.079 \; (0.084)$ & $9.9$ & $29.9$ &$65.5$ & $0.110 \pm 0.081 \; (0.094)$ & $97.0$  & $0.074 \pm 0.076 \; (0.093)$ \\																							
\hline																									
Regression	&	$76.6$ & $0.116 \pm 0.093 \; (0.123)$ & $17.5$ & $43.0$ &$61.2$ & $0.110 \pm 0.112 \; (0.068)$ & $95.4$  & $0.125 \pm 0.061 \; (0.123)$ \\																							
\hline \hline 																									
																									
\multirow{2}{*}{$\mathrm{S_{1+2}}$-AGT}	&	\multicolumn{4}{c|}{Entire Test Set}							&	\multicolumn{2}{c|}{Certain Test Studies (55\%)}							&	\multicolumn{2}{c|}{Uncertain Test Studies  (45\%)}							\\
\cline{2-9}																									
	&	acc (\%)	&	abs: mean $\pm$ std (median)	&	ece (\%)	&	mce (\%)	&	acc (\%)	&	abs: mean $\pm$ std (median)					&	acc (\%)	&	abs: mean $\pm$ std (median)					\\
\hline																									
CDF-Prob (Lap)	&	$81.9$ & $0.097 \pm 0.079 \; (0.077)$ & $20.8$ & $26.7$ &$69.9$ & $0.104 \pm 0.080 \; (0.085)$ & $\bf 96.6$  & $\bf 0.068 \pm 0.078 \; (0.089)$ \\ 																							
\hline																									
CDF-Prob	&	$\bf 82.9$ & $\bf 0.095 \pm 0.082 \; (0.073)$ & $22.5$ & $29.0$ &$\bf 72.3$ & $\bf 0.098 \pm 0.084 \; (0.075)$ & $95.9$  & $0.069 \pm 0.079 \; (0.090)$ \\																							
\hline																									
PDF-Prob	&	$81.4$ & $0.099 \pm 0.081 \; (0.079)$ & $13.5$ & $28.8$ &$70.0$ & $0.100 \pm 0.082 \; (0.080)$ & $95.2$  & $0.078 \pm 0.080 \; (0.097)$ \\																							
\hline																									
MC-Dropout	&	$81.3$ & $0.099 \pm 0.084 \; (0.079)$ & $22.1$ & $34.5$ &$70.4$ & $0.105 \pm 0.081 \; (0.087)$ & $94.5$  & $0.066 \pm 0.086 \; (0.092)$ \\																							
\hline																									
Regression	&	$81.6$ & $0.101 \pm 0.084 \; (0.083)$ & $22.8$ & $37.5$ &$70.2$ & $0.109 \pm 0.084 \; (0.089)$ & $95.4$  & $0.067 \pm 0.082 \; (0.092)$ \\																							
\hline \hline 																									
																									
\multirow{2}{*}{$\mathrm{S_{1+2}}$-SGT}	&	\multicolumn{4}{c|}{Entire Test Set}							&	\multicolumn{2}{c|}{Certain Test Studies (55\%)}							&	\multicolumn{2}{c|}{Uncertain Test Studies  (45\%)}							\\
\cline{2-9}																									
	&	acc (\%)	&	abs: mean $\pm$ std (median)	&	ece (\%)	&	mce (\%)	&	acc (\%)	&	abs: mean $\pm$ std (median)					&	acc (\%)	&	abs: mean $\pm$ std (median)					\\
\hline																									
CDF-Prob (Lap)	&	$80.7$ & $0.101 \pm 0.085 \; (0.080)$ & $17.5$ & $28.7$ &$69.9$ & $0.106 \pm 0.086 \; (0.085)$ & $94.1$  & $0.071 \pm 0.083 \; (0.095)$ \\																							
\hline																									
CDF-Prob	&	$\bf 82.2$ & $\bf 0.099 \pm 0.084 \; (0.079)$ & $23.2$ & $29.2$ &$\bf 71.3$ & $\bf 0.100 \pm 0.085 \; (0.079)$ & $\bf 95.4$  & $\bf 0.077 \pm 0.083 \; (0.097)$ \\ 																							
\hline																									
PDF-Prob	&	$79.9$ & $0.105 \pm 0.089 \; (0.081)$ & $22.7$ & $25.3$ &$68.4$ & $0.107 \pm 0.088 \; (0.083)$ & $94.1$  & $0.075 \pm 0.089 \; (0.102)$ \\																							
\hline																									
MC-Dropout	&	$76.8$ & $0.109 \pm 0.092 \; (0.105)$ & $12.2$ & $23.4$ &$63.1$ & $0.102 \pm 0.104 \; (0.061)$ & $93.6$  & $0.117 \pm 0.073 \; (0.118)$ \\																							
\hline																									
Regression	&	$78.4$ & $0.107 \pm 0.097 \; (0.103)$ & $14.0$ & $29.7$ &$66.1$ & $0.096 \pm 0.109 \; (0.044)$ & $93.4$  & $0.120 \pm 0.078 \; (0.119)$ \\ 																							
\hline \hline 																									
																									
\multirow{2}{*}{$\mathrm{S_1^{\mathrm{EX}}}$}	&	\multicolumn{4}{c|}{Entire Test Set}							&	\multicolumn{2}{c|}{Certain Test Studies (55\%)}							&	\multicolumn{2}{c|}{Uncertain Test Studies  (45\%)}							\\
\cline{2-9}																									
	&	acc (\%)	&	abs: mean $\pm$ std (median)	&	ece (\%)	&	mce (\%)	&	acc (\%)	&	abs: mean $\pm$ std (median)					&	acc (\%)	&	abs: mean $\pm$ std (median)					\\
\hline																									
CDF-Prob (Lap)	&	$\bf 82.8$ & $\bf 0.093 \pm 0.081 \; (0.072)$ & $15.5$ & $23.7$ &$\bf 71.0$ & $\bf 0.099 \pm 0.083 \; (0.080)$ & $\bf 97.3$  & $\bf 0.064 \pm 0.077 \; (0.086)$ \\																							
\hline																									
CDF-Prob	&	$81.1$ & $0.100 \pm 0.085 \; (0.079)$ & $22.5$ & $33.7$ &$68.5$ & $0.105 \pm 0.089 \; (0.084)$ & $96.3$  & $0.075 \pm 0.079 \; (0.094)$ \\																							
\hline																									
PDF-Prob	&	$80.3$ & $0.099 \pm 0.082 \; (0.079)$ & $19.7$ & $25.2$ &$68.2$ & $0.105 \pm 0.083 \; (0.088)$ & $95.2$  & $0.067 \pm 0.080 \; (0.092)$ \\																							
\hline																									
MC-Dropout	&	$78.1$ & $0.108 \pm 0.090 \; (0.107)$ & $12.6$ & $36.6$ &$64.2$ & $0.102 \pm 0.105 \; (0.059)$ & $95.0$  & $0.116 \pm 0.067 \; (0.115)$ \\																							
\hline																									
Regression	&	$77.1$ & $0.112 \pm 0.094 \; (0.123)$ & $17.1$ & $37.7$ &$62.9$ & $0.101 \pm 0.111 \; (0.054)$ & $94.5$  & $0.125 \pm 0.067 \; (0.126)$ \\																							
\hline																									
    \end{tabular}%
    }
   
    \label{tb:compare_methods_ensemble_performance}
\end{table*}

\section{The Semi-automated Ultrasound Beam Cropping System and Image Down-sizing}
\label{app:semi}
The main issue to crop the ultrasound beam from the DICOM images is that the beam shape and location varies from different machine model and maker, different settings such as depth.
The difficulty of using an image processing based method to crop the ultrasound beam is that some border region of the beam can be dark because there are no tissue present.
Therefore, these region are likely to be confused with background 2D pixels and being excluded despite these pixels are a part of the tissue and should be kept.
Our semi-automatic cropping method would require the user to manually trace the correct boundaries and save these beam shapes.
When a new image needs to be cropped, the automatic image processing method would propose a cropping shape and the system would compare it with the closest user-entered beam shape, then the closest user-entered shape (\ie, less than 5\% mismatched pixels) will be selected as the cropping shape.
If not, the system would ask the user to check and crop for this image, which makes this system semi-automatic.
In addition, users have the choice to let the system suppresses the questionable ones to be checked at the end when the automatic crop-able cases are all processed.

The DICOM B-mode images are either at $1024\times768$ or $800\times600$ resolution, and the ultrasound image beam are at the level of roughly 400 pixels and above in both dimensions.
As a result, the cropped raw ultrasound image beam is always larger than the defined network input size; hence can be down-sized to $120 \times 120$ with no zero-padding needed.

\begin{table*}[!htbp]
    \centering
    \caption{The one tail hypothesis test results computed by using two-sample $t$-test, testing whether the mean absolute error of a method (specified by the row name) is statistically significantly smaller than that of another method (specified by the column name). The symbol $\star$ marks the significance with 95\% confidence interval ($p < 0.05$) and $\circ$ marks the significance with 90\% confidence interval ($p < 0.1$).}
    \resizebox{\textwidth}{!}{%
    \begin{tabular}{|c|c||c|c||c|c||c|c|c|c|c||c|c|c|c|c||c|c|c|c|c|}

\hline \multicolumn{2}{|c||}{} &	\multicolumn{2}{c||}{$\mathrm{S_1}$}	&	\multicolumn{2}{c||}{$\mathrm{S_2}$}	&	\multicolumn{5}{c||}{$\mathrm{S_{1+2}}$-AGT}	&				\multicolumn{5}{c||}{$\mathrm{S_{1+2}}$-SGT}	&				\multicolumn{5}{c|}{$\mathrm{S_{1}^{EX}}$}					\\ \cline{3-21}
\multicolumn{2}{|c||}{}	& \rotatebox[origin=c]{90}{CDF-Prob}	& \rotatebox[origin=c]{90}{Regression}	& \rotatebox[origin=c]{90}{CDF-Prob}	& \rotatebox[origin=c]{90}{Regression}	& \rotatebox[origin=c]{90}{CDF-Prob (Lap)}	& \rotatebox[origin=c]{90}{CDF-Prob}	& \rotatebox[origin=c]{90}{PDF-Prob}	& \rotatebox[origin=c]{90}{MC-Dropout}	& \rotatebox[origin=c]{90}{Regression}	& \rotatebox[origin=c]{90}{CDF-Prob (Lap)}	& \rotatebox[origin=c]{90}{CDF-Prob}	& \rotatebox[origin=c]{90}{PDF-Prob}	& \rotatebox[origin=c]{90}{MC-Dropout}	& \rotatebox[origin=c]{90}{Regression}	& \rotatebox[origin=c]{90}{CDF-Prob (Lap)}	& \rotatebox[origin=c]{90}{CDF-Prob}	& \rotatebox[origin=c]{90}{PDF-Prob}	& \rotatebox[origin=c]{90}{MC-Dropout}	& \rotatebox[origin=c]{90}{Regression}	\\ \hline \hline
\multirow{2}{*}{\rotatebox[origin=c]{90}{$\mathrm{S_1}$}} & CDF-Prob	& 	& 	& 	& 	& 	& 	& 	& 	& 	& 	& 	& 	& 	& 	& 	& 	& 	& 	& 	\\ \cline{2-21}
& Regression	& 	& 	& 	& 	& 	& 	& 	& 	& 	& 	& 	& 	& 	& 	& 	& 	& 	& 	& 	\\ \hline \hline
\multirow{2}{*}{\rotatebox[origin=c]{90}{$\mathrm{S_2}$}} & CDF-Prob	& $\star$	& $\star$	& 	& $\star$	& 	& 	& 	& 	& 	& 	& 	& 	& $\star$	& 	& 	& 	& 	& $\circ$	& $\star$	\\ \cline{2-21}
& Regression	& 	& 	& 	& 	& 	& 	& 	& 	& 	& 	& 	& 	& 	& 	& 	& 	& 	& 	& 	\\ \hline \hline
\multirow{5}{*}{\rotatebox[origin=c]{90}{$\mathrm{S_{1+2}}$-AGT}} & CDF-Prob (Lap)	& $\star$	& $\star$	& $\circ$	& $\star$	& 	& 	& 	& 	& 	& 	& 	& $\star$	& $\star$	& $\star$	& 	& 	& 	& $\star$	& $\star$	\\ \cline{2-21}
& CDF-Prob	& $\star$	& $\star$	& $\star$	& $\star$	& 	& 	& 	& 	& $\star$	& $\star$	& 	& $\star$	& $\star$	& $\star$	& 	& $\circ$	& 	& $\star$	& $\star$	\\ \cline{2-21}
& PDF-Prob	& $\star$	& $\star$	& 	& $\star$	& 	& 	& 	& 	& 	& 	& 	& $\circ$	& $\star$	& $\star$	& 	& 	& 	& $\star$	& $\star$	\\ \cline{2-21}
& MC-Dropout	& $\star$	& $\star$	& 	& $\star$	& 	& 	& 	& 	& 	& 	& 	& $\circ$	& $\star$	& $\star$	& 	& 	& 	& $\star$	& $\star$	\\ \cline{2-21}
& Regression	& $\star$	& $\star$	& 	& $\star$	& 	& 	& 	& 	& 	& 	& 	& 	& $\star$	& $\circ$	& 	& 	& 	& $\star$	& $\star$	\\ \hline \hline
\multirow{5}{*}{\rotatebox[origin=c]{90}{$\mathrm{S_{1+2}}$-SGT}} & CDF-Prob (Lap)	& $\star$	& $\star$	& 	& $\star$	& 	& 	& 	& 	& 	& 	& 	& 	& $\star$	& $\circ$	& 	& 	& 	& $\star$	& $\star$	\\ \cline{2-21}
& CDF-Prob	& $\star$	& $\star$	& 	& $\star$	& 	& 	& 	& 	& 	& 	& 	& $\circ$	& $\star$	& $\star$	& 	& 	& 	& $\star$	& $\star$	\\ \cline{2-21}
& PDF-Prob	& $\star$	& $\star$	& 	& $\star$	& 	& 	& 	& 	& 	& 	& 	& 	& 	& 	& 	& 	& 	& 	& $\star$	\\ \cline{2-21}
& MC-Dropout	& 	& 	& 	& $\circ$	& 	& 	& 	& 	& 	& 	& 	& 	& 	& 	& 	& 	& 	& 	& 	\\ \cline{2-21}
& Regression	& $\star$	& $\circ$	& 	& $\star$	& 	& 	& 	& 	& 	& 	& 	& 	& 	& 	& 	& 	& 	& 	& $\circ$	\\ \hline \hline
\multirow{5}{*}{\rotatebox[origin=c]{90}{$\mathrm{S_{1}^{EX}}$}} &  CDF-Prob (Lap)	& $\star$	& $\star$	& $\star$	& $\star$	& 	& 	& $\circ$	& $\circ$	& $\star$	& $\star$	& $\circ$	& $\star$	& $\star$	& $\star$	& 	& $\star$	& $\circ$	& $\star$	& $\star$	\\ \cline{2-21}
& CDF-Prob	& $\star$	& $\star$	& 	& $\star$	& 	& 	& 	& 	& 	& 	& 	& 	& $\star$	& $\circ$	& 	& 	& 	& $\star$	& $\star$	\\ \cline{2-21}
& PDF-Prob	& $\star$	& $\star$	& 	& $\star$	& 	& 	& 	& 	& 	& 	& 	& $\circ$	& $\star$	& $\star$	& 	& 	& 	& $\star$	& $\star$	\\ \cline{2-21}
& MC-Dropout	& $\circ$	& 	& 	& $\star$	& 	& 	& 	& 	& 	& 	& 	& 	& 	& 	& 	& 	& 	& 	& 	\\ \cline{2-21}
& Regression	& 	& 	& 	& 	& 	& 	& 	& 	& 	& 	& 	& 	& 	& 	& 	& 	& 	& 	& 	\\ \hline

    \end{tabular}%
    }
    \label{tb:significance_table}
\end{table*}

\section{Numerical Results and Statistical Test Results}
\label{app:tables}

In Table~\ref{tb:compare_methods_ensemble_performance}, we include the numerical results for the compared methods illustrated in Fig.~\ref{fig:boxplot}.
The definition of the tested metrics can be found in Sec.~\ref{sec:exp_test_criteria} in the main article.
In Table~\ref{tb:significance_table}, we also include the one tail hypothesis test results computed by using two-sample $t$-test for the compared methods.

\section{The Details of the Toy Example}
\label{app:toy}
The top example shown in Fig.~\ref{fig:loss_comparison} describes a two-class (\{+1, -1\}) problem for an uncertain case where its observed labels are +1 or -1 with equal chance.
The x and y-axis labels $p_{+1}$ and $p_{-1}$ denote the inferred class probabilities from some learning model, and the z-axis shows the loss value computed by the cross-entropy function.
Since it is a two class problem, $p_{+1}$ and $p_{-1}$ sums to 1, where the loss values locate only on the light blue plane.

When the positive label +1 is observed, the loss value will be located on the red line $-\log(p_{+1})$ (simplified from the cross-entropy loss function), and the derivative (slope) points to right hand side.
When the negative label -1 is observed, the loss value will be located on the yellow line $-\log(p_{-1})$, which is opposite to the red line, and the derivative points to the opposite side.
When SGT is applied to the training, the randomness causing computed loss value jumping between the red and yellow lines; hence the direction of the derivative changes randomly.
However, at the region with drawn arrows, the yellow line always have a steeper slope (larger derivative) given the same x-y coordinates, which will eventually move the probabilities to middle where $p_{+1} = p_{-1} = 0.5$.

On the other hand, when AGT is used, the loss value will be located on the blue line $-\frac{1}{2}\log(p_{+1}) - \frac{1}{2}\log(p_{-1})$, assuming the probability of both labels being observed is known, resulting in a much more stable loss landscape with derivative always pointing to the middle.

\section{The Simulated Active Learning Environment Setting}
\label{app:active_learning}

By build upon the existing experiment setting described in Sec.~\ref{sec:experiments}, an active learning environment is simulated with the $\mathbf{A_{1}^{EX}}$ set acting as the expert cardiologist, and a learner simply inquires labels from it by nominating low confidence samples. 

All training samples are initially treated as unlabelled.
At the start of a training instance, the learner first nominates a random 25\% of the entire training pool of $\mathbf{A_{1}^{EX}}$ to get the corresponding labels, and the ``labelled'' samples are not to be returned to the pool.
For active learning experiments, we adopt the same training hyper-parameters. 
The most important modification is that the first 50 epochs are trained only on this 25\% of the data to allow a stable estimation of $\mathrm{conf}$, before inquiring labels.
Beyond the initial 50 epochs, for every two epochs, the learner first computes the confidence measure $\mathrm{conf}$ for the entire remaining ``unlabelled'' samples, and then requests the labels of a top 1.5\% samples (with the lowest $\mathrm{conf}$ values to $\mathbf{A_{1}^{EX}}$) as new training data (approx. $100$ to $150$ samples addition depending on the remaining volume).
At the end of the training, the learner will be able to see a total of approximate 50\% training data.

\end{appendices} 
  
\end{document}